\documentclass[10pt]{article}

\usepackage[T1]{fontenc}

\usepackage{stmaryrd}
\usepackage{amsthm}
\usepackage{amsmath}
\usepackage{amssymb}
\usepackage{mathrsfs}
\usepackage{soul}
\usepackage{pgf,tikz}
\usepackage{graphicx}
\usepackage{xcolor}
\usetikzlibrary{arrows}
\usepackage{listings}
\usepackage{hyperref}
\usepackage{cases}
\usepackage{mathabx}
\usepackage{algorithm2e}
\usepackage{algorithmic}
\usepackage{natbib}

\theoremstyle{plain}
\newtheorem{thm}{Theorem}[section]
\newtheorem{lem}[thm]{Lemma}
\newtheorem{prop}[thm]{Proposition}

\theoremstyle{definition}
\newtheorem{defn}{Definition}[section]

\theoremstyle{remark}
\newtheorem{rem}{Remark}

\def\Jac{\mathfrak{J}}

\def\R{\mathbb{R}}
\def\dim{d}
\def\samples{n}

\def\Rnm{\R^{\dim\times\samples}}
\DeclareMathOperator*{\argmin}{argmin}

\newcommand\smallmat[1]{\left[\begin{smallmatrix}#1\end{smallmatrix}\right]}
\def\bnu{\boldsymbol{\nu}}
\DeclareMathOperator{\mine}{\min_\varepsilon}

\def\bs{\mathbf{s}}
\def\bt{\mathbf{t}}
\def\bT{\mathbf{T}}
\def\bx{\mathbf{x}}
\def\by{\mathbf{y}}
\def\bz{\mathbf{z}}
\def\ba{\mathbf{a}}
\def\bb{\mathbf{b}}
\def\bq{\mathbf{q}}

\def\bg{\mathbf{g}}
\def\bw{\mathbf{w}}

\def\bu{\mathbf{u}}
\def\bv{\mathbf{v}}
\def\bbf{\mathbf{f}}

\def\diag{\mathrm{diag}}

\def\R{\mathbb{R}}

\def\Xcal{\mathcal{X}}
\def\Tcal{\mathcal{T}}
\def\Ucal{\mathcal{U}}
\def\Vcal{\mathcal{V}}
\def\Zcal{\mathcal{Z}}

\def\ones{\mathbf{1}}

\def\mine{\text{min}_\varepsilon}

\def\KR{\widetilde{R}}
\def\Kquantile{\widetilde{S}}

\newcommand{\dotp}[2]{\ensuremath{\langle #1 , #2\,\rangle}}

\def\R{\mathbb{R}}

\def\ones{\mathbf{1}}

\def\mine{\text{min}_\varepsilon}

\def\KR{\widetilde{R}}
\def\Kquantile{\widetilde{S}}
\def\quan{\widetilde{T}}

\def\Plm{P^\varepsilon_{\ell_-}}
\def\Plp{P^\varepsilon_{\ell_+}}
\def\Pe{P^\varepsilon_{\star}}

\newcommand{\eqdef}{:=}

\newcommand{\opSR}[2]{\KR_\varepsilon\left(#1;#2\right)}
\newcommand{\opSquantile}[2]{\Kquantile_\varepsilon\left(#1;#2\right)}
\newcommand{\opSquan}[3]{\quan_{#3}\left(#1;#2\right)}

\usepackage{microtype}
\usepackage{graphicx}
\usepackage{subfigure}
\usepackage{booktabs} 

\usepackage{hyperref}

\usepackage[accepted]{icml2020}

\icmltitlerunning{Quantile Normalization for Matrix Factorization}

\begin{document}

\twocolumn[
\icmltitle{Supervised Quantile Normalization for Low-rank Matrix Approximation}



\icmlsetsymbol{equal}{*}

\begin{icmlauthorlist}
\icmlauthor{Marco Cuturi}{goo}
\icmlauthor{Olivier Teboul}{goo}
\icmlauthor{Jonathan Niles-Weed}{nyu}
\icmlauthor{Jean-Philippe Vert}{goo}
\end{icmlauthorlist}

\icmlaffiliation{goo}{Google Research, Brain Team}
\icmlaffiliation{nyu}{New York University}

\icmlcorrespondingauthor{Marco Cuturi}{cuturi@google.com}

\icmlkeywords{quantile normalization, matrix factorization, optimal transport}
\vskip 0.3in
]



\printAffiliationsAndNotice{}  

\begin{abstract}
Low rank matrix factorization is a fundamental building block in machine learning, used for instance to summarize gene expression profile data or word-document counts. To be robust to outliers and differences in scale across features, a matrix factorization step is usually preceded by ad-hoc feature normalization steps, such as \texttt{tf-idf} scaling or data whitening. We propose in this work to learn these normalization operators jointly with the factorization itself. More precisely, given a $d\times n$ matrix $X$ of $d$ features measured on $n$ individuals, we propose to learn the parameters of quantile normalization operators that can operate row-wise on the values of $X$ and/or of its factorization $UV$  to improve the quality of the low-rank representation of $X$ itself. This optimization is facilitated by the introduction of a new differentiable quantile normalization operator built using optimal transport, providing new results on top of existing work by (Cuturi et al. 2019). We demonstrate the applicability of these techniques on synthetic and genomics datasets.
\end{abstract}

\section{Introduction}
The vast majority of machine learning problems start with a matrix $X\in\Rnm$ of measurements that keeps track of $\dim$ features measured on $\samples$ individuals. An important way to summarize the information contained in $X$ is to find a low-rank matrix factorization, namely two matrices $U$ and $V$ of sizes $d\times k$ and $k\times n$ such that $UV\approx X$, as quantified in a relevant matrix norm. While this problem is known to boil down to the truncated singular value decomposition of $X$ when the norm is Euclidean, the two most recent decades have succeeded in producing extremely useful variations on that problem, handling for instance the cases in which the entries of $X$ are non-negative~\citep{lee1999learning,hofmann2001unsupervised,fevotte2011algorithms}, binary~\citep{NIPS2013_4860} or even describe rank values~\citep{le2015rank}; considering various forms of sparse priors on the factors themselves~\citep{d2005direct,mairal2010online,jenatton2011proximal,witten2009penalized}; and extending these problems to cases where the matrices are incomplete~\citep{5197422,candes2009exact}.

\textbf{Low Rank Approximations} Let $\R^{\dim\times\samples}_k$ be the set of $d\times n$ matrices of rank $k$. Choosing a divergence $\Delta: \Xcal \times \Xcal\rightarrow \R$ defined on a subset of matrices $\Xcal\subset \Rnm$, and $\Zcal_k \subset \Xcal\cap \R^{\dim\times\samples}_k$, one can introduce the operator
$$
\Pi_k(X) := \argmin_{Z\in\Zcal_k} \Delta(X,Z).
$$ 
While a very large literature has focused on considering various divergences $\Delta$, such as Frobenius, KL~\cite{lee1999learning}, Beta-divergences~\cite{fevotte2011algorithms} or Wasserstein~\cite{pmlr-v51-rolet16}; and sets $\Zcal_k$ (sparse, non-negative), an important practical limitation of these approaches is that they perform well if the values described in $X$ have a distribution that is somewhat shared across features: Because the discrepancy $\Delta$ is usually additive, the loss $\Delta$ can be impacted by differences in ranges. This problem is addressed by ``massaging'' the entries of $X$ first, notably through ad-hoc normalization schemes carried out feature-by-feature, such as taking logarithms for gene expression data~\citep{Risso2018general} or using \texttt{tf-idf} schemes for text data, before feeding this modified matrix $\tilde{X}$ to the projector $\Pi_k$. 
\begin{figure*}[hthp]\label{fig:intro}
\includegraphics[width=\textwidth]{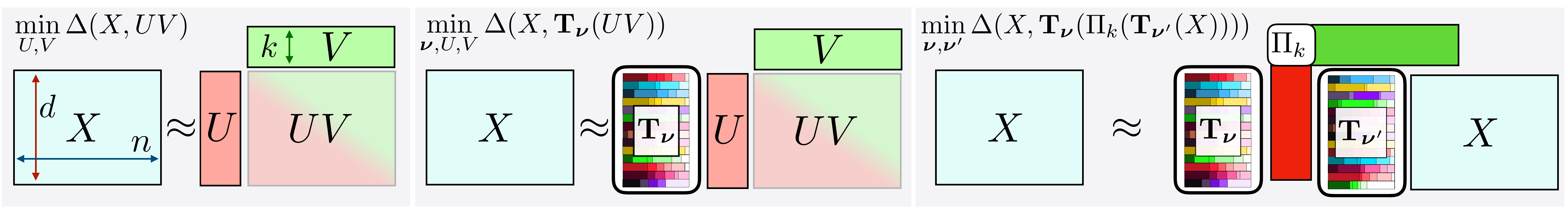}
\vskip-.4cm
\caption{\textit{(Left)} Schematic description of classic matrix factorization, with rank $k$ factor matrices $U,V$. 
\textit{(Middle)} QMF, the approach we propose to re-normalize the values of each single row $i$ of a factorization $UV$ a quantile renormalization operator defined per feature, using a target measure $\nu_i$. Each measure $\nu_i$ is described by a probability vector $\bb_i$ of size $m$ and a vector $\bq_i$ of $m$ quantiles (increasing values) at those levels. $\bnu$ stores all of these distributions. \textit{(Right)} Assuming one has access to a (differentiable) projector $\Pi_k$, we can require additionally that $X$ be itself the quantile normalization of a reconstruction of itself, after another quantile normalization operation. Here $\bT_{\bnu'}$ acts as a ``deflating'' mapping done to facilitate factorization, ``reinflated'' by $\bT_{\bnu}$ to yield the best possible reconstruction.}\end{figure*}

\textbf{Increasing Feature rescaling.} We propose in this paper to automatically learn such a renormalization, rather than leave to the user the arduous choice of selecting a suboptimal method. We also claim that we can gain interpretability by then finding out which features seem to be inflated / deflated to improve factorization. Our approach allows for an increasing map, defined for each of the $\dim$ features, to be applied to all $n$ values of the row of a matrix, either prior to and/or after the factorization step. The benefit of increasing maps is that they preserve the relative order of samples, which is an important point for interpretation. Denoting by $\mathcal{T}$ the set of increasing 
maps from $\R$ to $\R$, we consider a family of $\dim$ such maps $\mathbf{T}=(T_1,\dots,T_{\dim}) \in \mathcal{T}^\dim$, to which we associate (using the same symbol) an operator applying each map $T_i$ to the corresponding row of a $d\times n$ matrix $W$:
$$
\mathbf{T}(W) = [T_i(W_{ij})]_{ij} \in \Rnm.
$$
Our goal is to find one (or possibly two) maps $\bT$ that can work hand-in-hand with matrix factorization to minimize reconstruction error. We consider first \textit{quantile matrix factorization} (QMF), which minimizes $\Delta(X,\bT(UV))$ jointly in $U,V$ and $\bT$. We introduce an alternative approach involving two maps $\bT,\bT'$, to minimize $\Delta(X,\bT(\Pi_k(\bT'(X)))$ (QMFQ), see Fig.~\ref{fig:intro}.

\textbf{Scaling with Quantile Normalization.} To define and optimize families of maps $T_i$, we need to represent monotonic invertible functions in a parametric form that is amenable to optimization. While several approaches have been proposed recently to parameterize such maps \citep[and references therein]{NIPS2017_6891,NIPS2019_8433,JMLR:v17:15-243}, we propose here a new approach that can be conveniently optimized with respect to both the parameters of $T$ as well as its inputs, with the added benefit that one can control \textit{exactly} the row-wise distributions of the outputs of $\bT$. This can be useful for instance to enforce similarities between the values taken jointly across one or more several rows, or to ``pin'' these values to lie in a segment, as we do in our experiments. In order to reach that property, we parameterize each map $T_i$ as a \textit{quantile normalization} operator w.r.t to a target measure, written as $T_{\nu_i}$. Differentiation is achieved by extending the toolbox of~\citet{cuturi2019differentiable} to include soft-quantile normalisation operators.


\textbf{Contributions} Our contributions are two-fold: \textbf{(i)} After introducing recent tools of~\citet{cuturi2019differentiable}, we improve them in three ways in \S\ref{sec:back}: we add to their operators a new differentiable quantile normalization operator; we prove the monotonicity of all these operators, putting these tools on a sound footing; and we derive the implicit differentiation of these operators, rather than unrolling Sinkhorn iterations.
\textbf{(ii)} We introduce low-rank factorization models in \S\ref{sec:models} that employ this soft-quantile normalization layer, and propose various algorithmic approaches to train them (including stochastic schemes), either relying on implicit or explicit factorizations as in Fig.\ref{fig:intro}. We test these approaches in \S\ref{sec:exp} on synthetic datasets and on real multiomics cancer data.
\section{Differentiable Quantile Normalization using Optimal Transport}\label{sec:back}
We recall the approach proposed recently by~\citet{cuturi2019differentiable} to view ranking and sorting problems as optimal transport problems that can be turned into differentiable operators through regularization. We then proceed with three contributions: \textit{\textbf{(i)}} We extend their operators to define a \textit{quantile normalization operator} $\quan_{\varepsilon, \bb, \bq}$  which takes an array of weighted values $\bx$ and modifies them so that these values now follow a given target quantile distribution as described by $\bb$ and $\bq$. The parameter $\varepsilon>0$ is a smoothing parameter to ensure differentiability. \textit{\textbf{(ii)}} We prove the monotonicity of Sinkhorn-ranks, Sinkhorn-sort and of the newly introduced Sinkhorn-quantile normalization operators. This is an important result that was missing from~\citet{cuturi2019differentiable} (informally, proving the the curves in the middle plot of the Figure 2 can never cross) and that is also crucial to ground to  work on solid footing, since we can thus guarantee that our functions  $\quan_{\varepsilon, \bb, \bq}$ are indeed increasing and therefore conserve ranks. \textit{\textbf{(iii)}} We introduce an implicit differentiation scheme of the solutions of regularized OT, which offers an interesting alternative to the automatic differentiation of Sinkhorn iterations that was put forward by ~\citet{cuturi2019differentiable}.

\textbf{Notation.} We denote by $\Sigma_n = \{\bu\in\R^n_+ | \bu^T\ones_n=1\}$ the set of $n$-dimensional probability vectors. For any vector, $\bx=(x_1,\dots)$, we write $\overline{\bx}$ for its cumulative-sum vector, namely the vector with entries $[\sum_{j\leq i} x_j]_i$. When applied on a matrix $R$, the same operator $\overline{R}$ denotes the cumsum operation applied \textit{row-wise}. A function $c:\R\times\R\rightarrow \R$ is submodular if it is twice differentiable and $\partial^2 c/\partial x\partial y<0$. For two probability vectors $\ba,\bb$ of size $n$ and $m$, we write $U(\ba,\bb)=\{P\in\R^{n\times m}_+ |  P\ones_m=\ba, P^T\ones_n=\bb\}$ for the transportation polytope. Operations on matrices are to be understood elementwise, and we use $\circ$ for the elementwise product between matrices or vectors.

\subsection{Background: soft-ranking/sorting using OT}
Suppose one is given an array $\bx=(x_1,\dots,x_n)$ of $n$ numbers, weighted by a positive probability vector $\ba=(a_1,\dots,a_n)$ of the same size. The idea of~\citet{cuturi2019differentiable} is to consider an auxiliary vector $\by=(y_1,\dots,y_m)$ of ordered values---typically the regular grid of $m$ values in $[0,1]$---to form a cost matrix $C_{\bx\by}:=[c(x_i,y_j)]_{ij}$, with $c$ submodular. Along with probability vector $\bb=(b_1,\dots,b_m)$ for $\by$, one defines then a (primal) regularized OT problem:
\begin{equation}\label{eq:rot}
P_\star^\varepsilon :=  \argmin_{P\in U(\ba,\bb)} \dotp{P}{C_{\bx\by}}-\varepsilon H(P)\,,\tag{P-RegOT}
\end{equation}
where $H(P) = -\sum_{i,j} P_{ij} \left( \log P_{ij} - 1\right)$ denotes $P$'s entropy. This regularized OT problem has a factorized solution~\cite{CuturiSinkhorn} $P_\star^\varepsilon$ which can be written as $\diag(\bu)K\diag(\bv)$, where $K=\exp(-C_{\bx\by}/\varepsilon)$ and $\bu\in\mathbf{R}^n$ and $\bv\in\mathbf{R}^m$ are fixed points of the Sinkhorn iteration. 
~\citet{cuturi2019differentiable} proposed the following smoothed ranking and sorting operators
$$
\begin{aligned}
	\opSR{\ba,\bx}{\bb,\by} &:= n \ba^{-1}\circ (\Plp \overline{\bb})\in[0,n]^n,\\ 
	 \opSquantile{\ba,\bx}{\bb,\by} & := \, \bb^{-1}\circ  ({\Plm}^T \bx)\in\R^{m}.
\end{aligned}$$
obtained after running the Sinkhorn algorithm, and writing $$\Plp=\diag(\bu_\ell) K \diag(\bv_\ell), \Plm=\diag(\bu_{\ell-1}) K \diag(\bv_\ell).$$ 
The Sinkhorn algorithm is described in simplified form in Alg.\ref{algo-sink}. In practice the number of iterations $\ell$ can be set dynamically, to enforce convergence.

\textbf{Modification to Guarantee Marginals.} There is a small but important modification we have done to these operators, compared to~\citet{cuturi2019differentiable}: We consider in the definition of $\Kquantile_\varepsilon$ the scaling $\bu_{\ell-1}$ and not the scaling $\bu_{\ell}$. This is done in order to take advantage of the the fact that after any iteration on $\bv$ in algorithm Alg.\ref{algo-sink}, the transport matrix estimate $\diag(\bu_{i-1})K\diag(\bv_i)$ has column-sums exactly equal to $\bb$ (but row-sums not necessarily equal to $\ba$), whereas $\diag(\bu_{i})K\diag(\bv_i)$ has the opposite property (equality of row-sums to $\ba$ is ensured, but not of column-sums to $\bb$). This modification ensures that the operators $\KR_\varepsilon$ and $\Kquantile_\varepsilon$ effectively apply row-stochastic kernels to their inputs, so each of the entries of $\KR_\varepsilon$ and $\Kquantile$ are convex combinations of $\bx$ and $n \overline{\bb}$. These modifications are particularly important small $\ell$.
\begin{algorithm}[H]
\SetAlgoLined
\textbf{Inputs:} $\ba,\bb,\bx,\by,\varepsilon,c$

$C_{\bx\by} \gets [c(x_i, y_j)]_{ij}$,\;
$K \gets e^{-C_{\bx\by}/\varepsilon},\;\bu=\ones_n$\;

\For{$i=1,\dots,\ell$}{
    $\bv_i\gets \bb/K^T\bu_{i-1},\;\bu_{i}\gets \ba/K\bv_{i}$
  }
\KwResult{$\bu_{\ell},\bv_{\ell},\bu_{\ell-1},K$}
\caption{Sinkhorn with $\ell$ iterations}\label{algo-sink}
\end{algorithm}

\subsection{Differentiable quantile normalization}\label{subsec:diffquan}
In the continuous world, a quantile normalization operator maps values distributed according to a measure $\mu$ to values described in a measure $\nu$. That map is increasing~\citep[\S2]{SantambrogioBook}, and is the composition of maps $Q_\nu\circ F_\mu$: one computes first the CDF of the input value $x$ w.r.t. $\mu$, and then outputs the quantile of $\nu$ at that level. The challenge of that transformation when instantiated on two discrete measures $\nu=\sum_j b_j \delta_{q_j}$ and $\mu=\sum_i a_i \delta_{x_i}$, is that it is not differentiable because quantile and CDF functions $Q_\nu$ and $F_\mu$ are staircase-like functions, and the output of their composition, requiring ranking, sorting and lookup tables is not continuous. We adopt now a discrete perspective leveraging the operators defined above to define a differentiable quantile normalization operator that takes the values of $\bx$ (with weights $\ba$) as inputs, computes their soft-transport to a target measure $(\bb,\by)$ where $\by$ is an arbitrary increasing sequence in $\R^m$, and then computes a convex combination of the quantiles as described in $\bq$:

\begin{defn}[Soft-Quantile Normalization Operators] For any increasing vector $\bq\in\R^m$ paired with a vector of weights $\bb\in\Sigma_m$, the following operator denotes the quantile renormalization with respect to $\bq$ of the values in $\bx$ weighted by $\ba$:
\begin{equation}\label{eq:quan}\opSquan{\ba,\bx}{\by}{\varepsilon,\bb,\bq} \eqdef \ba^{-1}\circ (\Plp\bq) \in\R^n.\end{equation}
\end{defn}
\begin{rem}
Since $\diag(\ba^{-1}) \Plp$ is row-stochastic, the entries of $\widetilde{T}$ are convex combinations of entries of $\bq$.
As we show below, the entries of this vector have the same relative ordering as the entries of $\bx$.\end{rem}
\begin{rem}As the regularization level $\varepsilon\rightarrow 0$ and if $m=n$, $\ba=\bb=\ones/n$, then $P_\star^\varepsilon$ converges to the sorting permutation matrix of $x$, $\KR_\varepsilon$ to the vector of ranks, and $\quan_\epsilon$
to the usual quantile normalization operator (the entries of $\widetilde{T}_\epsilon$ are exactly those of $\bq$ reindexed to agree with the arg-sort of $\bx$), see~\citet{morvan2017supervised}.\end{rem}

\subsection{Monotonicity of Sinkhorn Operators}
To be consistent as smoothed ranking, sorting and quantile normalization operators, $\KR_\varepsilon, \Kquantile_\varepsilon$ and $\quan_{\varepsilon,\bb,\bq}$ should possess basic monotonicity properties.
The sorted vector $\opSquantile{\ba,\bx}{\bb,\by}$ should be non-decreasing, and if $x_i \leq x_{i'}$ then the $i$th entry of $\KR_\varepsilon$ (respectively, $\widetilde{T}_\varepsilon$) should be smaller than its $i'$th entry.
As the following proposition shows, the smoothed ranking and sorting operators proposed here enjoy both of these properties, for any number of Sinkhorn iterations.
\begin{prop}\label{prop:monotone}
For any $\ell \geq 0$ and any submodular cost $c$, the following relations hold:
\begin{align*}
    j \leq j' & \implies [\opSquantile{\ba,\bx}{\bb,\by}]_j \leq [\opSquantile{\ba,\bx}{\bb,\by}]_{j'} \\
    x_i \leq x_{i'} & \implies [\opSR{\ba, \bx}{\bb, \by}]_i \leq [\opSR{\ba, \bx}{\bb, \by}]_{i'} \\
    x_i \leq x_{i'} & \implies [\opSquan{\ba,\bx}{\by}{\varepsilon,\bb,\bq}]_i \leq [\opSquan{\ba,\bx}{\by}{\varepsilon,\bb,\bq}]_{i'}\,.
\end{align*}
\end{prop}
The proof is given in the supplementary, and relies on stochastic monotonicity of the rows of the iterations of the Sinkhorn algorithm, thanks to the crucial assumption that $c$ is submodular~\cite{chiappori2017multi}. A remarkable feature of this result is that it holds regardless of the number of iterations $\ell$. 

\subsection{Implicit Differentiation of Sinkhorn Operators}\label{subsec:implicit}
\citet{cuturi2019differentiable} proposes to use a direct automatic differentiation of Sinkhorn iterations to obtain differentiability of the transports $\Plm$ and $\Plp$ that appear within $\KR_\varepsilon, \Kquantile_\varepsilon, \quan_{\varepsilon,\bb,\bq}$. Because storing all Sinkhorn iterations is required to use automatic differentiation, the RAM cost of this approach is heavy, totalling at least $O(\ell n+m + nm)$, notably when $\varepsilon$ is small, in which case $\ell$ can be typically several hundred. A possible workaround would be to use faster regularized OT solvers. However, most of the approaches investigated recently to speed up the Sinkhorn iterations yield non-differentiable computational graphs, since they involve conditional choices~\cite{dvurechensky2018computational} or are not easy to parallelize~\cite{altschuler2017near}. We have tried modified iterations~\citep{thibault2017overrelaxed,schmitzer2016stabilized} but they still prevent the use of these operators in larger scale settings. We propose to bypass this issue using implicit differentiation. 

\textbf{Implicit Differentiation for Backpropagation} Since our goal is to backpropagate through the transport we need a fast algorithm to apply the \textit{transpose} of the Jacobian map of $\quan_{\varepsilon,\bb,\bq}$ only w.r.t. inputs $\bx,\bb,\bq$ (for reasons that will become clear in the next section).  A variant of the computations below appears in~\cite{NIPS2018_7827}, which we complement in several ways: Their goal was to compute the gradient of $\dotp{\Pe}{C_{\bx\by}}$ w.r.t. $\bb$ only, their method uses a Cholesky factorization of a $(n+m)\times(n+m)$ matrix, while we do away with this step using a Schur complement and generalize derivations to also inputs $\bx$.
Similar computations have also been carried out to prove statistical results by~\citet{klatt2018empirical}.

\textbf{Differentiating the OT Matrix} The main challenge when computing the transpose-Jacobians of $\KR_\varepsilon, \Kquantile_\varepsilon$ and $\quan_{\varepsilon,\bb,\bq}$, is to differentiate the optimal transport matrix $\Pe$ w.r.t any of the relevant inputs (both $\Plm$ and $\Plp$ are assumed to have converged to the solution $\Pe$, which we must assume to do implicit calculus). Since $\Pe$ does not change with $\bq$ as apparent from~\eqref{eq:quan} \citep[a fact highlighted by][]{morvan2017supervised}, the transpose-Jacobian of $\quan_{\varepsilon,\bb,\bq}$ w.r.t. $\bq$ is simply the map $H\rightarrow (H\circ\Pe)^T\ones_n$. The challenge is therefore to provide a fast way to apply the transpose-Jacobians of $\Pe$ to any matrix $H$ of size $n\times m$ w.r.t. $\bx$ and $\bb$.

Given $H\in\R^{n\times m}$, and a variable $u$ that is either $\bx$ or $\bb$, we seek a linear operator $(\Jac_u\Pe)^T$ such that, assuming $\Pe(u+du)-\Pe(du) \approx \Jac_u\Pe(du)$, one has for any $H$ that $\dotp{H}{\Jac_u\Pe(du)}=\dotp{(\Jac_u\Pe)^T H}{du)}$. We parameterize first $\Pe$ as the solution of a dual regularized OT problem.

\textbf{Dual formulation} As discussed by \citet[\S4]{COTFNT},~\eqref{eq:rot} is equivalent to the dual problem below, where we use the notation $\bbf\oplus\bg:=\bbf\ones_m^T + \ones_n\bg^T$ for the tensor addition of these two vectors,
\begin{equation}\label{eq:dregot}\tag{D-RegOT}
\!\!\!\max_{\bbf \in\R^n, \bg\in\R^m} \!\!\bbf^T \ba + \bg^T \bb - \varepsilon \ones_n^T e^{\frac{\bbf\oplus\bg-C(u)}{\varepsilon}}\ones_m.
\end{equation}
When $\bbf$ and $\bg$ are optimal, one has $\Pe=e^{\frac{\bbf\oplus\bg-C(u)}{\varepsilon}}$.

\textbf{Variations in $\Pe$.} Assuming all other variables fixed,
\begin{multline*}\Pe(u+du)-\Pe(u) \approx \Jac_u\Pe(du) \\= \frac{1}{\varepsilon}\Pe(u)\circ \left(\Jac_u\bbf(du)\ones_m^T+\ones_n \Jac_u\bg(du)^T - \Jac_uC(du)\right)\\
=\frac{1}{\varepsilon} \diag(\Jac_u\bbf(du))\Pe(u)+\Pe \diag(\Jac_u\bg(du)) \\
- \Pe(u)\circ \Jac_uC(du).\end{multline*}
Using an arbitrary $H$, one recovers that
\begin{multline*} \dotp{H}{\Jac_u\Pe(du)}= \tfrac{1}{\varepsilon} \dotp{(\Jac_u\bbf)^T (H\circ\Pe)\ones_m}{du} \\+ \tfrac{1}{\varepsilon} \dotp{(\Jac_u\bg)^T (H^T\circ{\Pe}^T)\ones_n}{du} 
\\
-\tfrac{1}{\varepsilon}\dotp{(H\circ P\circ\Jac_uC^T)\ones_m}{du}.\end{multline*}

Three (transposed) Jacobians are therefore needed w.r.t. $u$, those of $\Jac_u\bbf$, $\Jac_u\bg$ and $\Jac_uC$. Note that $\Jac_{\bb}C=\mathbf{0}_{(n\times m)\times m}$ and we will assume $(\Jac_\bx C)^T$ can be accessed. For instance when $c=(\cdot-\cdot)^2$, $(\Jac_\bx C)^T= \Delta:=2(\by\ones_n^T - \ones_m\bx^T)$. Jacobians of $\bbf(u)$ and $\bg(u)$ require more work.

\textbf{Variations in $\bbf,\bg$.} We use the $\mine$ operator to simplify equations, defined on matrices of arbitrary size as
$$\text{For}\, A \in \R^{p\times q}, \mine(A)=-\varepsilon \log((e^A)^T\ones_q)\in\R^p.$$
 For a vector $\bz\in\R^{n+m}$ we write $\bz_f$ and $\bz_g$ for the vectors of its first $n$ and last $m$ entries respectively, such that $\bz=\smallmat{\bz_f\\\bz_g}$. The first order conditions for $\bbf,\bg$, boil down to 
$$\!\tau\left(\smallmat{\bbf\\\bg}\right) =0, \text{where } \tau(\bz)\! :=\! \smallmat{\mine(C_{\bx\by}- \bz_f\oplus\bz_g) +\varepsilon \log \ba\\ \mine(C^T_{\bx\by}- \bz_g\oplus\bz_f) +\varepsilon \log \bb}.$$

A sequence of computations (provided in the supplement) yields that the Jacobian of $\tau$ is a $n\times m$ block matrix,
\begin{equation*}
   \Jac_\bz \tau(\bx,\bz)= - \begin{bmatrix} I_n & M_1 \\ M_2 & I_m  \\ \end{bmatrix} \,,
\end{equation*}
where we have written $M=e^{(\bz_f\oplus\bz_g-C_{\bx\by})/\varepsilon}$ to define $M_1=\diag(1/M\ones_m)M$ and $M_2=\diag(1/M^T\ones_n)M^T$. Note that in the case where $\bz$ coincides with a solution $\smallmat{\bbf\\\bg}$ to \eqref{eq:dregot}, $M$ is the optimal transportation plan. $M_1$ and $M_2$ can therefore be interpreted as Markov kernel (row-stochastic) matrices. Using the matrix inversion theorem we obtain that the inverse \textit{transposed} Jacobian is
\begin{equation*}
\!- \Jac_\bz \tau(\bz)^{-T}\! =\! \begin{bmatrix} I_n + M_2^T S^{-1} M_1^T & - M_2^T S^{-1}  \\\!\! - S^{-1}M_1^T  &\!\! S^{-1}  \\ \end{bmatrix},
\end{equation*}
with the transpose-Schur complement $S = I_m-M_1^TM_2^T$.

\textbf{Differentiation w.r.t $\bx$ or $\bb$}
The implicit mechanism linking variable $u$ (where $u$ is still either $\bx$ or $\bb$) with $\bbf(u)$ and $\bg(u)$ is given by the implicit function theorem (here, instantiated in its transpose form) which states that, at optimality (here we overload $\tau$ to also consider values $\bx$ or $\bb$ as inputs to the first order equation for simplicity), 
$$\left(\Jac_u \smallmat{\bbf(u)\\\bg(u)}\right)^T=  - (\Jac_u \tau(u))^{T}  (\Jac_\bz \tau(\bz))^{-T},$$
With a few computations we have for $u$ equal to $\bx$ or $\bb$,
$$(\Jac_\bx \tau)(\bx) d\bx= \smallmat{d\bx \circ (M1\circ\Delta)\ones_m \\ (M_2\circ\Delta^T)d\bx}, (\Jac_\bb \tau(\bb)) d\bb= \smallmat{\mathbf{0}_n \\ \varepsilon d\bb/\bb},$$
and therefore 
$$\begin{aligned}(\Jac_\bx \tau)^T(d\bz) &= \bz_f\circ (M_1\circ\Delta)\ones_m + (M_2^T\circ\Delta)d\bz_g,\\ (\Jac_\bb \tau)^T (d\bz)&= \bb/\varepsilon\circ d\bz_g.\end{aligned}$$
Applying these results backwards, we recover Alg.~\ref{algo-lsesink}, which provides all ``custom'' gradients needed to incorporate Sinkhorn operators in end-to-end differentiable pipelines.

\begin{algorithm}[H]
\SetAlgoLined
\textbf{Inputs:} $\ba,\bb,\bx,\by,\varepsilon,c,\rho>0,H\in\R^{n\times m}$

$C \gets [c(x_i,y_j)]_{ij}$,\;
$\Delta \gets (\Jac_\bx C)^T.$

\Repeat{$\|e^{-\tfrac{1}{\varepsilon}(C^T-\bg\oplus\bbf)}\ones_n-\bb\|<\rho$}{
    $\bg \gets \varepsilon \log \bb + \mine(C^T-\bg\oplus\bbf) + \bg$
    
    $\bbf \gets \varepsilon \log \ba + \mine(C-\bbf\oplus\bg) + \bbf$
  }

$P \gets e^{-\tfrac{1}{\varepsilon}(C^T-\bg\oplus\bbf)}, \bbf \gets \bbf - \bbf[0], \bg \gets \bg + \bbf[0]$

$M_1 \gets \diag(1/P\ones_m) P, M_1[0,:]=\mathbf{0}_{1\times m}$ 

$M_2 \gets \diag(1/P^T\ones_n) P^T$

$\bz_f \gets (H\circ P)\ones_m,\, \bz_g \gets (H\circ P)^T\ones_n$

$\bw_g \gets S^{-1} (M_1^T\bz_f-\bz_g).$

$\bz_f \gets \bz_f + M_2^T \bw_g, \bz_g \gets -\bw_g$

$\Jac_\bx^T H = -\frac{1}{\varepsilon}(H\circ P\circ\Delta)\ones_m + \bz_f \circ (M_1\circ\Delta)\ones_m + (M_2^T\circ\Delta)\bz_g$

$\Jac_\bb^T H = \bb\circ \bz_g /\varepsilon$

\KwResult{$P,\Jac_\bx^T H,\Jac_\bb^T H$}
\caption{Sinkhorn and Jacobians Transpose Evaluations}\label{algo-lsesink}
\end{algorithm}

\begin{rem}Alg.\ref{algo-lsesink} above contains two extra steps not appearing in our presentation. Those consist in setting to $0$ the first entry of $\bbf$ (and offsetting all other entries) and deleting the first row of $M_1$. This modification is due to the fact, also noticed by ~\citet{NIPS2018_7827}, that $\bbf$ and $\bg$ are determined up to a constant. Pinning the first variable of $\bbf$ to $0$ helps lift this indeterminacy, and slightly modifies $M_1$ by removing its first row, ensuring $S$ is invertible.
\end{rem}
\begin{rem}The implicit approach outlined here is particularly well suited to the case where $m\ll n$, since the linear system to be solved is of size $m\times m$ and dominates the cost of the final iterations outlined in Alg.~\ref{algo-lsesink}. As can be expected, we do observe in practice that the execution time of Alg.~\ref{algo-lsesink} is roughly half that of the backprop approach used in~\cite{cuturi2019differentiable}. The biggest improvement is of course in terms of memory. All of the experiments done next exploit this approach, and were stable numerically.
\end{rem}
\begin{figure}
    \centering
    \includegraphics[width=.4\textwidth]{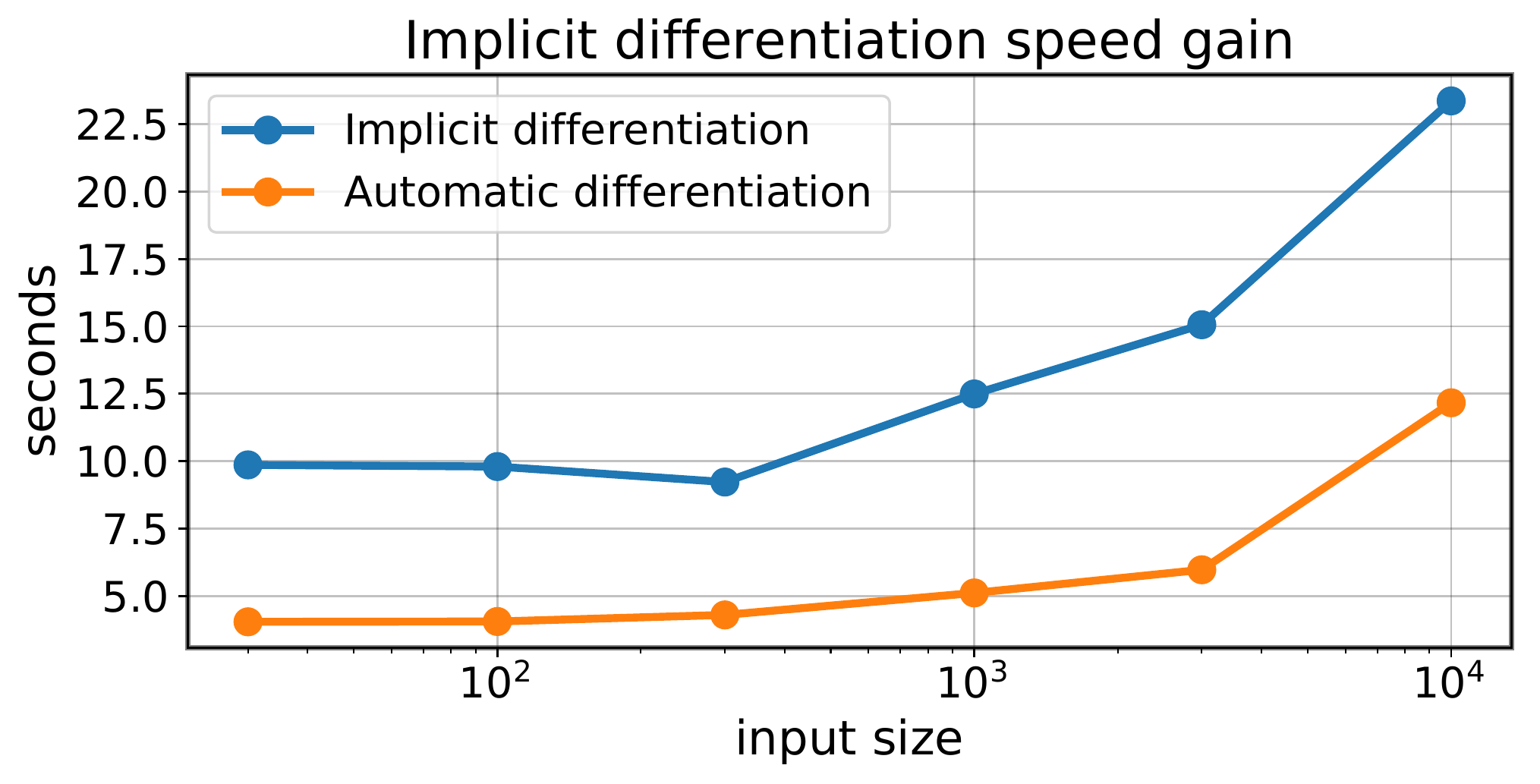}
    \vskip-.5cm\caption{Comparison of raw computation times (including tensorflow instantiation) of transposed Jacobians using AD and the implicit approach outlined in Algo.\ref{algo-lsesink}. Execution carried n batches, with 10 averages, $m=10$ and $\varepsilon=10^{-2}$}
    \label{fig:my_label}
    \vskip-.4cm
\end{figure}
\section{Matrix Factorization using Quantile Renormalization}\label{sec:models}
 We explore in this section the problem of finding a good low-rank approximation $Z\in \R^{d\times n}_k$ to a data matrix $X\in\R^{d\times n}$ using the tools introduced in \S\ref{sec:back}. 
 
\subsection{Scaling and Factorization Models} The most standard way to express the matrix factorization problem is to search for a low-rank matrix $Z=UV$ that directly approximates $X$, where $Z$ is the product of a tall and slim matrix $U\in\Ucal_k\subset\R^{\dim\times k}$ and a short and fat matrix $V\in\Vcal_k\subset\R^{k\times n}$.
That approximation is measured in terms of $\Delta(X,UV)$ for some divergence $\Delta$ (Figure~\ref{fig:intro}, left). As mentioned in \S1, we propose first to consider a family $\bT$ of row-wise monotonic transform, such that $X$ is approximately equal to $\bT(UV)$ for a product of factors $UV$ (Figure~\ref{fig:intro}, middle). This suggests to consider the matrix-factorization using supervised quantile normalization:
\begin{equation}\label{eq:minUV}\tag{QMF}
    \min_{\bT\in\Tcal^d, U\in\Ucal_k, V\in\Vcal_k} \Delta(X, \bT(UV))\,.
\end{equation}

We propose a second formulation that provides a more fined grained control on the low-rank factorization itself, and which requires the simultaneous optimization of two transformations $\bT$ and $\bT'$ to balance two different goals: \textit{(i)} the transformation of $X$ under $\bT'$ should facilitate obtaining a low-rank approximation, namely by minimizing the gap between $\bT'(X)$ and $\Pi_k(\bT'(X))$; 
\textit{(ii)} the low-rank approximation should allow, up to another transformation $\bT$, the approximate recovery of $X$, i.e., $X \approx \bT(\Pi_k(\bT'(X)))$.
These two goals---ease of factorization without sacrificing the ability to reconstruct---can be modelled through the following bi-level optimization problem ($\Pi_k$ being itself a projection, the solution of an optimization)
\begin{equation}\label{eq:minPik}\tag{QMFQ}
    \min_{\bT,\bT'\in\Tcal^d} \Delta(X, \bT(\Pi_k(\bT'(X))))\,.
\end{equation}
Note that if there is a transformation $\bT_X\in\Tcal^d$ such that $\bT_X(X)$ is exactly low-rank, then taking $\bT'=\bT_X$, $\bT = \bT_X^{-1}$ solves (\ref{eq:minUV}) and (\ref{eq:minPik}). In the general case, though, there is no reason why we should have $\bT' = \bT^{-1}$ at optimum.
Note that (\ref{eq:minPik}) is harder to minimize than (\ref{eq:minUV}), since any solution $\bT, \bT'$ of (\ref{eq:minPik}) leads to a feasible point of (\ref{eq:minUV}) with the same value (by taking $UV = \Pi_k(\bT'(X))$), so the optimal value of (\ref{eq:minUV}) is at least as low as that of (\ref{eq:minPik}). We now describe, building on the different results in \S2, our approach to parameterize the families of maps $\bT, \bT'$.

\subsection{Feature Scaling using Quantile Normalization.}
 Recall that given two distributions $\mu$ and $\nu$ over $\R$, a quantile normalization operator $T_{\mu\rightarrow\nu}:\R\rightarrow\R$ takes samples distributed according to $\mu$ and applies a non-decreasing transformation such that these samples, after this transformation, are distributed according to $\nu$. Writing $F_\mu:\R\rightarrow (0,1)$ for the cumulative distribution function (CDF) of $\mu$ and $Q_{\nu}:(0,1)\rightarrow \R$ for the quantile function of $\nu$, such a non-decreasing map can be written as
$$T_{\mu\rightarrow\nu} := Q_\nu \circ F_\mu.$$ 
Notice therefore that, using classic identities relating the CDF and quantile functions of a measure, one has
$$T_{\mu\rightarrow\nu}^{-1}= F^{-1}_\mu \circ Q^{-1}_\nu= Q_\mu \circ F_\nu =T_{\nu\rightarrow\mu},$$

\textbf{From data and variables to measures.} In both ~\eqref{eq:minUV} and ~\eqref{eq:minPik} problems, the input measure $\mu$ will be the discrete measure of values extracted from the row of a matrix $W$, where $W$ will be either $X$ directly (as in QMFQ), or a low-rank reconstruction, either in explicit $UV$ or implicit $\Pi_k$ form. We parameterize the output measure $\nu$ as a discrete measure of finite support. To solve~\eqref{eq:minUV} or ~\eqref{eq:minPik} in $\bT$, one would then require that the outputs of $T_{\mu\rightarrow \nu}$ be differentiable according to \textit{both} the input measures (notably when applied to a reconstruction) and the parameterized output measure $\nu$. The differentiation w.r.t to quantiles themselves was investigated by~\citet{morvan2017supervised}. The differentiability w.r.t. inputs can be obtained using the soft-quantile normalization operator introduced in \S\ref{subsec:diffquan}. Note that our definition also has the added flexibility, compared to~\citet{morvan2017supervised}, of introducing weighted quantiles with parameter $\bb$ (this is equivalent to defining exactly the levels $\overline{\bb}$ to which these quantiles correspond).

\subsection{Row-wise soft-quantile transformation}

\textbf{Input measures} $\boldsymbol{\mu}=(\mu_1,\dots,\mu_d).$
$\mu_i$ is the distribution of values of the $i$-th feature of a matrix, written here $W$. $W$ can be the original data matrix $X$ or, more to the point for optimization, its explicit ($UV$) or implicit ($\Pi_k(X)$) reconstructions. The measure for feature $i$ is therefore $\mu_i(W) := \frac{1}{n} \sum_{j=1}^n \delta_{W_{ij}}$.

\textbf{Target measures} $\boldsymbol{\nu}=(\nu_1,\dots,\nu_d)$.
We store the values of vectors $\bb_i$ and $\bq_i$ row-wise in matrices $B,Q$ to obtain
$\nu_i(B,Q) := \frac{1}{n} \sum_{j=1}^m B_{ij} \delta_{Q_{ij}},$
where $B,Q\in\R^{d\times m}$, $B$ is row-stochastic and $Q$ is row increasing. Notice $B$ and $Q$ are simultaneously pictured in Fig.\ref{fig:intro} underneath the labels $\bT_{\bnu}$ : the colors varying from dark to light denote increasing values for $Q$ row-wise, whereas the varying sizes of buckets in each row stand for probability weights $B$. Here, parameter $m$ effectively controls the complexity / size of $\nu_i$. We argue that in most cases, the budget of target quantiles $m$ should be much lower that $n$, as discussed as well by~\citet{cuturi2019differentiable}. In applications where $n$ is a few hundreds, we find that choosing small $m$, such as 8 or 16, works very well (See  Fig.~\ref{fig:learnedq}).

\textbf{From discountinuous to everywhere differentiable.} Given an arbitrary sorted sequence $\by=(y_1,\dots,y_m)$, our approach relies on the following identity, valid only when $n=m$ and $\bb=\ba=\tfrac{\ones_n}{n}$ (and therefore $B=\ones_{d\times n}/n$), and for each $i\leq d$, $\bw=W_{i\cdot},\bq=Q_{i\cdot}$
$$\opSquan{\ba,\bw}{\by}{\varepsilon,\bb,\bq_i} \underset{\varepsilon\rightarrow 0}{\rightarrow} T_{\mu_i\rightarrow\nu_i}(\bw).$$
The main interest in using the expression on the left is that, either through back-propagation or implicit differentiation as proposed in \S\ref{subsec:implicit}, the $\quan_{\varepsilon,\bb,\bq_i}$ operator is differentiable in all of its inputs and parameters (crucially $W,B,Q$) here as soon as $\varepsilon>0$, while $T_{\mu_i\rightarrow\nu_i}$ is not.

\textbf{Row-stochasticity of $B$.} The constraints that $B$ is row-wise stochastic can be taken into account by introducing precursors under the action of a soft-max. More specifically, we will write $B$ as the row-wise softmax of a precursor matrix of $F\in\R^{d\times m}$,
$$B(F) := \sigma(F) = \diag(1/(e^F\ones_m)\,e^{F}.$$

\textbf{Monotonicity and Range of $Q$.}  The constraint that $Q$ is increasing can be enforced by considering cumulative sums of non-negative values, possibly offset by a constant. In its most direct form, notably when deflating $X$ to obtain a matrix easy to factorize as in \eqref{eq:minPik}, $Q$ can be cast as the row-wise cumulative sum of the exponentials of an arbitrary precursor matrix $R\in\R^{d\times m}$,
$$Q'(R) := \overline{e^{R}},$$
where the cumsum operator is applied row-wise. When a quantile operator is carried out to inflate back the results of low-rank factorizations $UV$ or $\Pi_k$, we can directly ``pin'' the quantiles to lie in a range of values known beforheand. Indeed, since the goal is to reconstruct a known data matrix $X$, one can set those segments to be $[s_i,t_i]$ where $s_i$ and $t_i$ are the minimums and maximums of row $i$. Therefore, for a slightly slimmer matrix $R\in\R^{d\times (m-1)}$, and storing the ranges of values in $\bs=(s_1,\dots,s_d)$ and $\bt=(t_1,\dots,t_d)$, we recover the following map to define suitable quantiles from a precursor $R$,
$$Q(R,\bs,\bt) := \diag(\bt-\bs)[\mathbf{0}_{d},\overline{\sigma(R)}]+\bs\ones^T_m,$$
therefore recovering increasing quantiles that are pinned down to lie exactly in the desired ranges.

\subsection{Quantile Matrix Factorization}
We treat the first problem outlined in \eqref{eq:minUV} as an optimization problem with two precursor variables $F,R\in\R^{d\times m}$ that characterize target measures through quantile values $Q(R)$ and probability weights $B(F)$. We assume that $\Delta$ is separable along rows, and use the following notations for space: Given two precursor matrices $F$ and $R$, the map $\tilde{\bT}_{\varepsilon,F,R}$ applies to each row $i$ the soft quantile operator defined
using weights $\bb_i = B(F)_{i\cdot}$ and quantiles $\bq_i = Q'(R)_{i\cdot}$.

$$\min_{\substack{R\in\R^{n\times {m-1}}\\
F\in\R^{n\times m}\\ U\in\R^{d\times k},V\in\R^{k\times n}}}\sum_{i} \Delta\left(X,\tilde{\bT}_{\varepsilon,F,R}(UV)\right).$$
Note that the main computational effort here consists in applying $d$ quantile normalization operators. When suitable, we therefore use mini-batch sampling on the $d$ features to perform SGD on all parameters. When used on non-negative matrix factorization problems, as demonstrated in \S4, we also parameterize $U,V$ as exponential maps of precursor matrices of the same size to enfore non-negativity.

\subsection{Quantiles Matrix Factorization Quantiles}
The optimization problem outlined in \eqref{eq:minPik} is a bilevel programming problem, and therefore less scalable than QMF. We consider it nonetheless because of its interest as a modeling tool: The result of QMFQ can be used to normalize first a new incoming point, project it on the dictionary $U$ resulting from $\Pi_k(\bT'(X))$, and then project it back using $\bT$. To optimize this bilevel problem, we consider here the case in which $\Pi_k$, the (approximate) projection operator can be computed with an accesss to an operator computing the transpose of the Jacobian applied to an input matrix. This is notably the case when using SVD, with the analytic formulas provided for truncated SVD~\citep[Thm. 25]{feppon2018geometric}, or by unrolling a fixed point iteration, such as the multiplicative updates proposed in \citep{lee1999learning} to minimize the Kullback-Leibler loss between two non-negative matrices. We consider here the latter approach to consider for $F,F',R'\in\R^{d\times m}$ and $R\in\R^{d\times (m-1)}$.

$$\min_{\substack{R\in\R^{n\times {m-1}}\\
F,F',R'\in\R^{n\times m}}}\sum_{i} \Delta\left(X, \tilde{\bT}_{\varepsilon,F,R}(\Pi_k(\tilde{\bT}_{\varepsilon,F',R'}(X))))\right).
$$

\section{Experiments}\label{sec:exp}
In all experiments reported here, we set $\varepsilon$ and learning rates to $0.01$.
\begin{figure*}[hthp!]
    \centering
    \includegraphics[width=\textwidth]{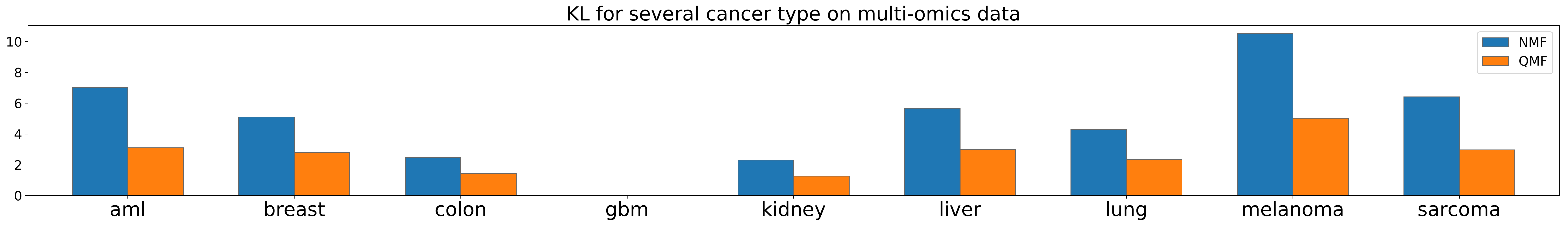}
    \vskip-.4cm\caption{KL losses for various cancer datasets as described in ~\citep{chalise17}, in which dimensions are a few hundreds patients for $n$ and $d~\approx 11,500$ per dataset. Here $QMF$ is computed with small batch size 64 for QMF along the $\dim$ dimension and $m=16$.}
    \label{fig:cancerloss}
\end{figure*}
\textbf{Toy illustrations} 
We consider in this section the following dimensions: $\dim=160$, $n=80$, $k=8$. We generate two ground truth factors $U_*$ and $V_*$ randomly, $U_*$ is a table of i.i.d Poisson realizations with parameter $\lambda=2$, whereas each column of $V_*$ is drawn according to a Dirichlet prior with parameters $\alpha=1/2$. We then apply a ``ground truth'' quantile normalization to these entries, $\bT_{0,F_*,R_*}(U_*V_*)$, where the precursors $R$ are sampled as a standard Gaussian multivariate distribution, and $F_*$ is a vector of zeros of size $m_*=n$ (using here standard quantile renormalization, not regularized). We then run NMF, QMF and QMFQ using the same $k=8$.

\begin{figure}[hthp]
    \centering
    \includegraphics[width=.4\textwidth]{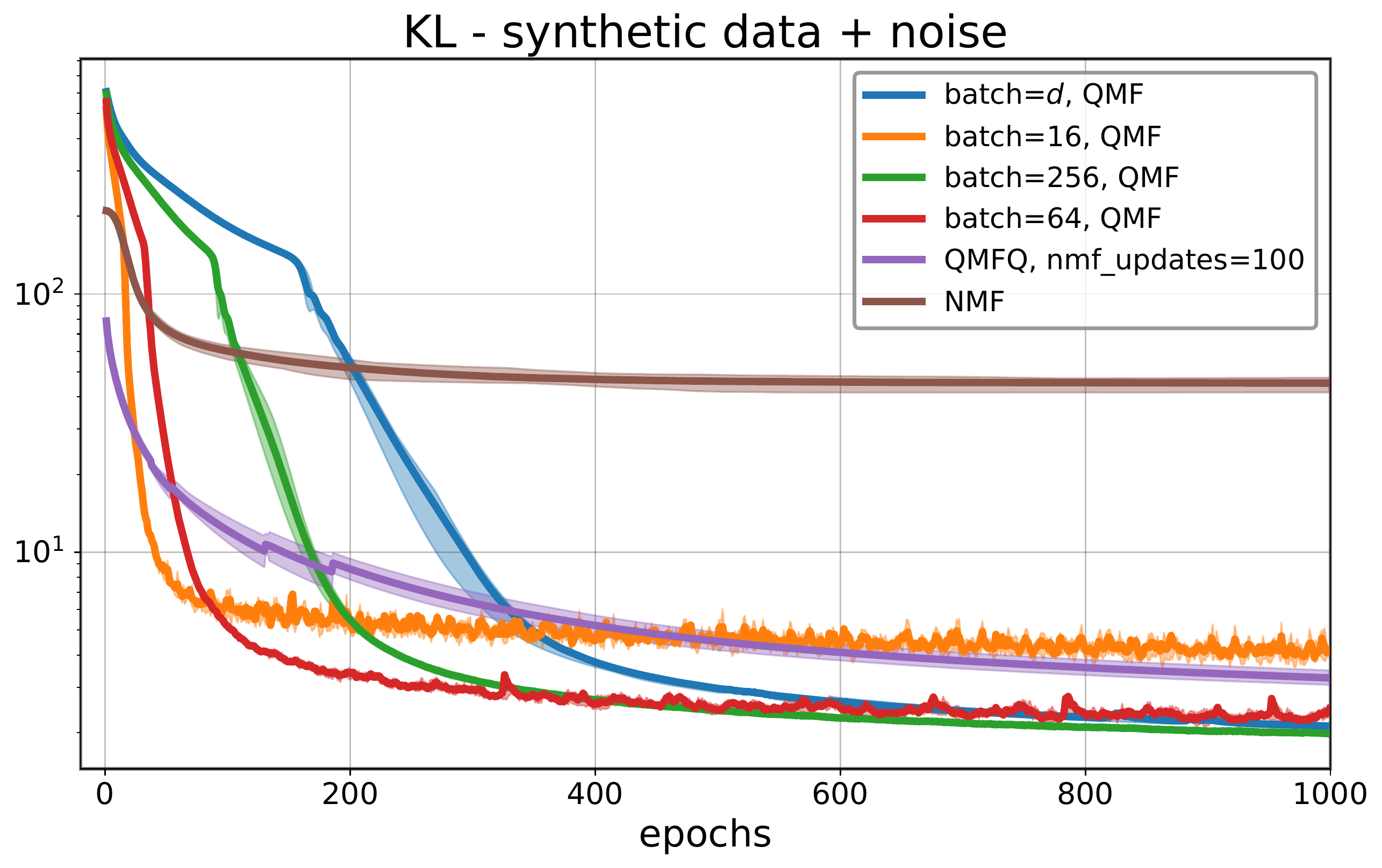}
    \vskip-.5cm
    \caption{KL decrease in a synthetic $\bT_*(UV)+ \max(10\,\mathcal{N},0)$ model with noise. As can be expected, the size of the batch size for QMF influences early/late convergence.}
    \label{fig:my_label4}
\end{figure}

\begin{figure}[hthp]
    \centering
    \includegraphics[width=.45\textwidth]{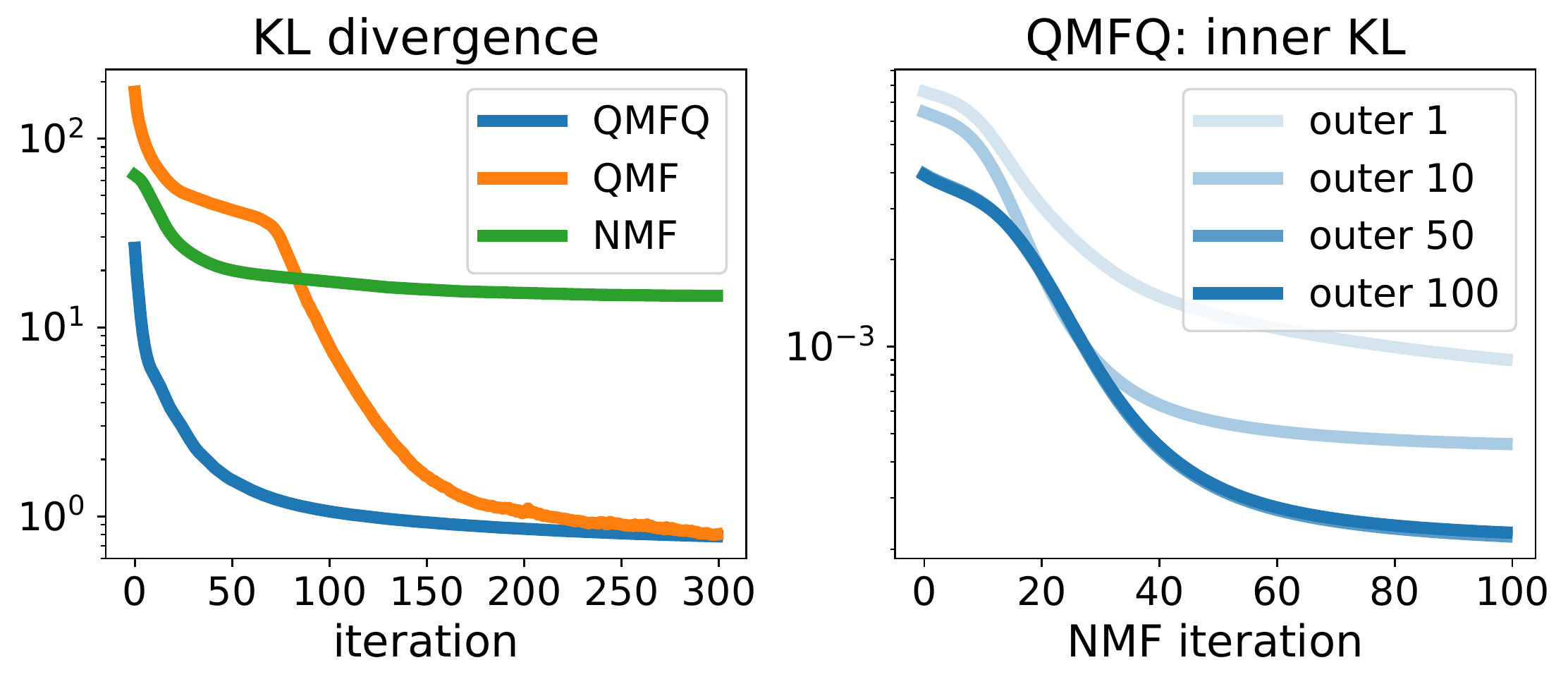}
    \vskip-.5cm
    \caption{(left) Decrease of the KL loss across iterations for all three methods. The NMF quickly saturates (the ground truth is indeed \textit{not} low rank), while QNMF and QNMFQ quickly reach almost perfect reconstruction, despite going through a quantile normalization step that only has a budget of $m=8$ values. Note that the plot only describes outer loop iterations. In that sense QMFQ is far slower than QMF or NMF, since it requires a 100 iterations of NMF as an inner loop to correctly approximate $\Pi_k(\bT'(X))$. (right) decrease of these inner loops as a function of the outer loop. As expected, the KL decreases faster as the algorithm progresses. Note that the KL scales on both plots are not comparable, since the right KL is computed using values taking different ranges of values (see also Fig.~\ref{fig:learnedq})}
    \label{fig:qnmf}
\end{figure}

\begin{figure}[hthp]
    \centering
    \includegraphics[width=0.45\textwidth]{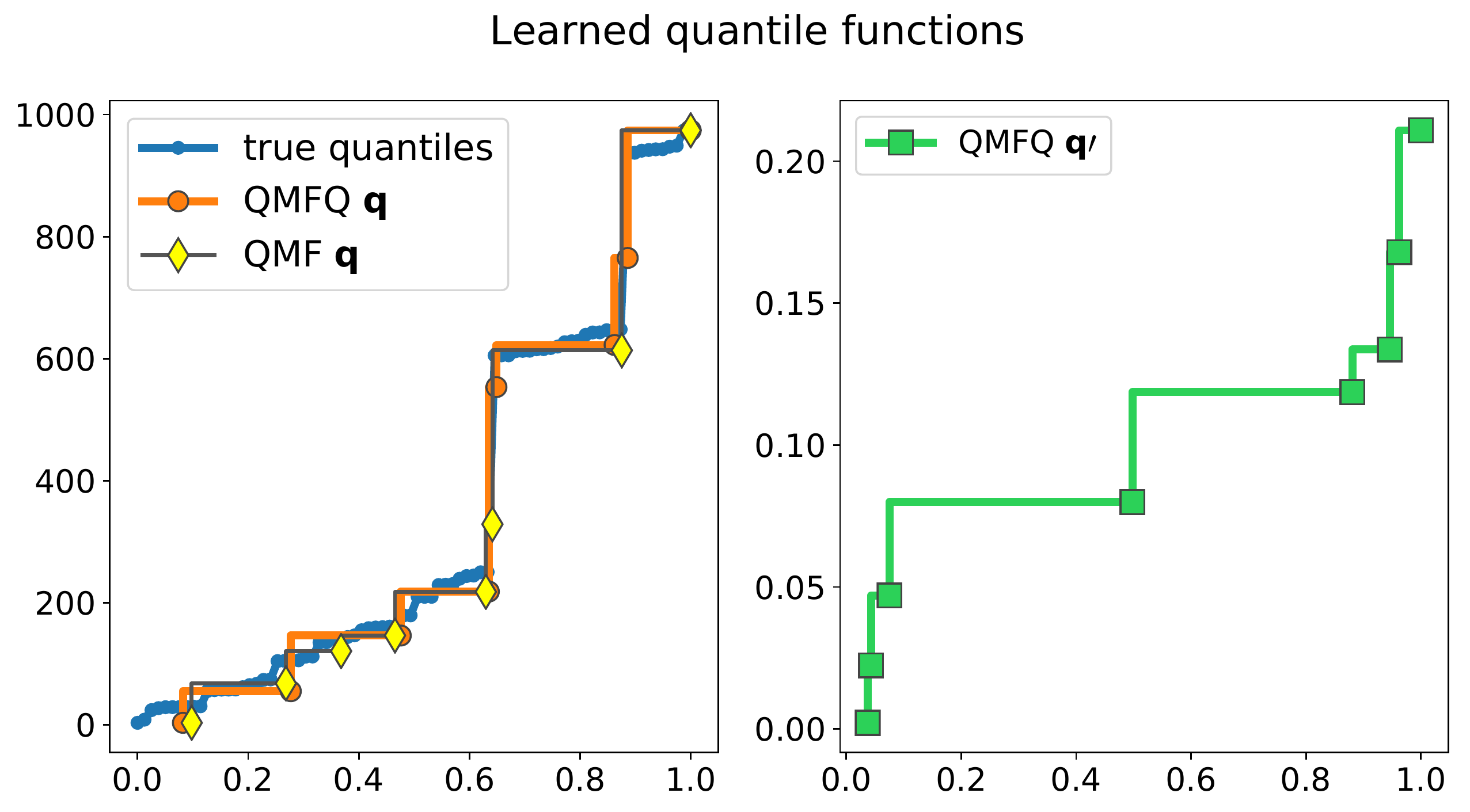}
    \vskip-.3cm\caption{(left) In blue, the true quantile distribution used to modify the values of the first row of the ground truth matrix $U_*V_*$. Both QMFQ and QMF methods are able to recover almost exactly lighter representations (here $m=8$) of the true quantile distributions. (right) In addition, QMFQ also produces a transformation able to deflate the values of $X$ so that they can be easily factorized. Note the difference in ranges ($y$-axis) between the values of $\bq$ and $\bq'$.}
    \label{fig:learnedq}
\end{figure}

The factors $U,V$ used in NMF, QMF and each inner evaluation of $\Pi_k$ in QMFQ are initialized with random uniform values (to retain consistency across outer iterations, the seed of QMFQ is always the same). We plot in Fig.~\ref{fig:qnmf} the KL divergence of these three different methods. We plot in Fig.~\ref{fig:learnedq} the two quantile distributions quantiles learned by QMFQ for the first feature, as well as the learned quantile for QMF.

\textbf{Error bars on larger experiments}
We consider the following dimensions, $\dim=500$, $n=256$, $k=10$, and run the algorithms across various setups, including mini-batches for QMF, various inner loop iterations for QMFQ and various values for targets $m$. We monitor the KL decrease averaged over 8 repeats of the data generation process outlined above (quantile normalization of $U_*,V_*$) to which we add a truncated Gaussian noise (non-negative values) of standard deviation 10. All of our results (see supplementary for more exhaustive explorations of the parameter space) agree with intuition and show the robustness of the two approaches presented here, and are summerized in Fig.~\ref{fig:my_label4}.

\textbf{Genomics.}
As an illustration on real-world data we consider the problem of multiomics data integration, a domain where NMF has been shown to be a relevant approach to capture low-rank representations of cancer patients using multiple omics datasets~\citep{chalise17}. Following the recent benchmark of \citet{Cantini2020.01.14.905760}, we collected from The Cancer Genome Atlas (TCGA) three types of genomic data (gene expression, miRNA expression and methylation) for thousands of cancer samples from 9 cancer types, and compare a standard NMF to QMF in their ability to find a good low-rank approximation of the concatenated genomic matrices. Figure~\ref{fig:cancerloss} confirms that on all cancers, QMF finds a factorization much closer to the original data than NMF does. We provide more detailed results in Annex B.2.

\begin{figure*}[htb!]
    \centering
    \includegraphics[height=6cm]{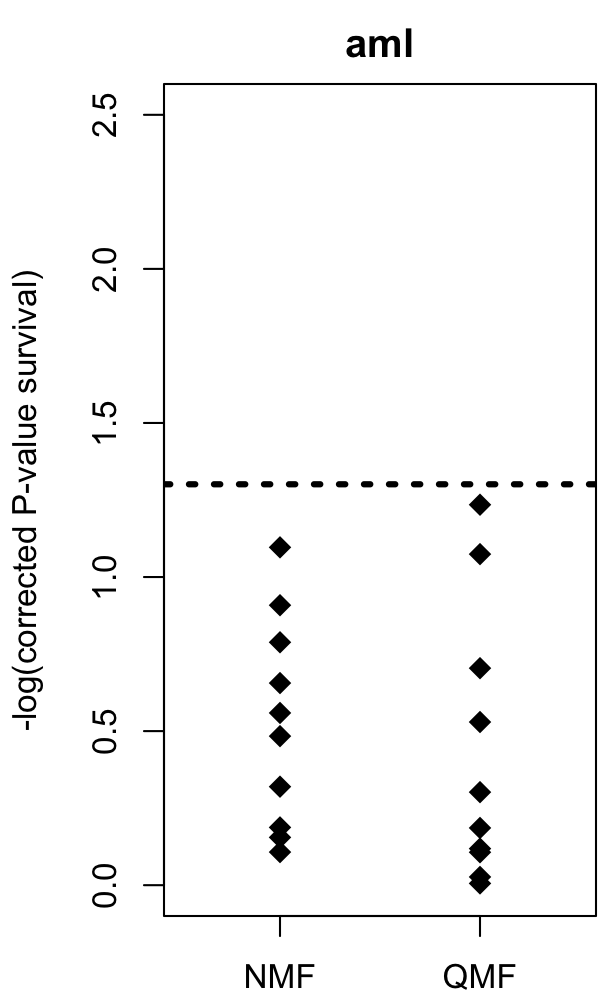}
    \includegraphics[height=6cm]{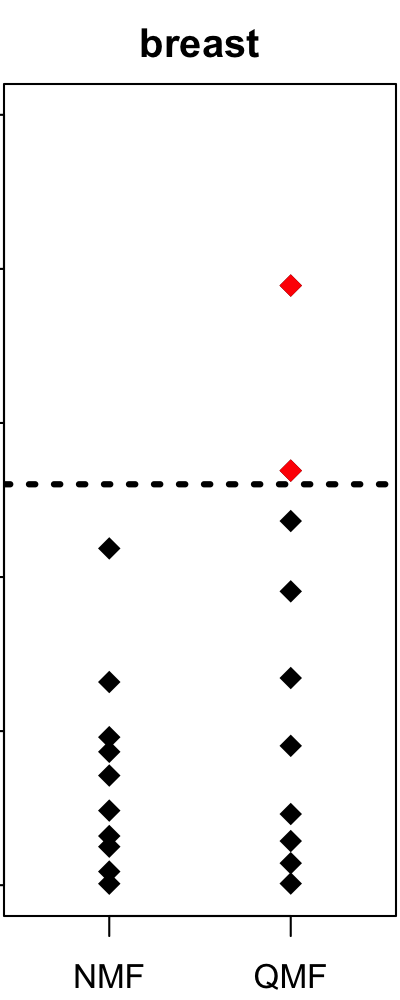}
    \includegraphics[height=6cm]{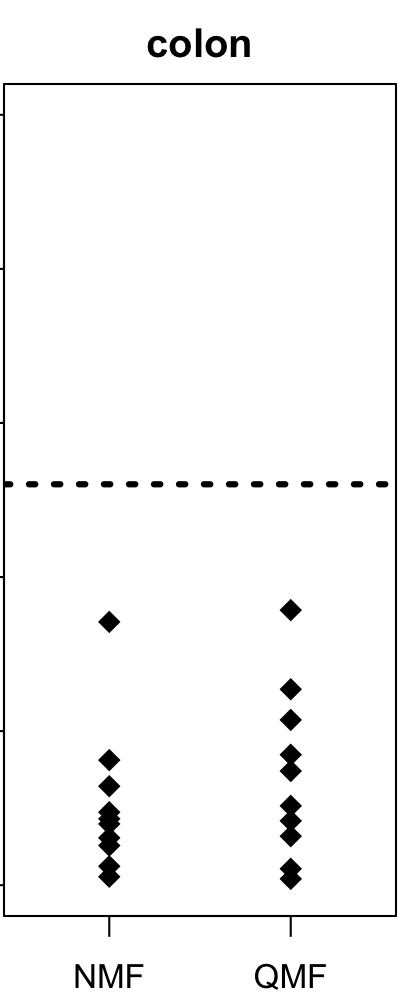}
    \includegraphics[height=6cm]{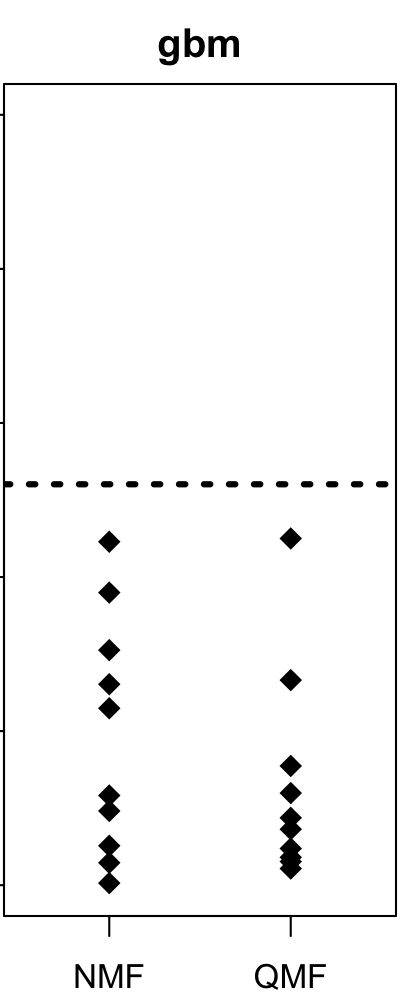}
    \includegraphics[height=6cm]{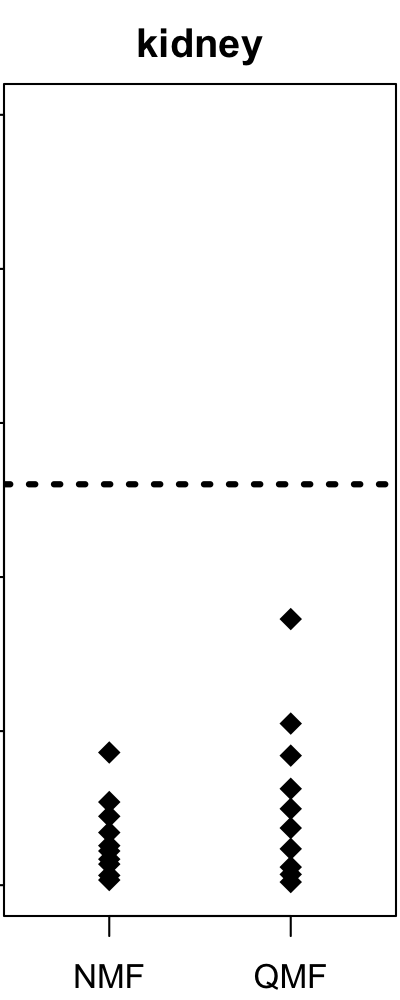}\\
    \includegraphics[height=6cm]{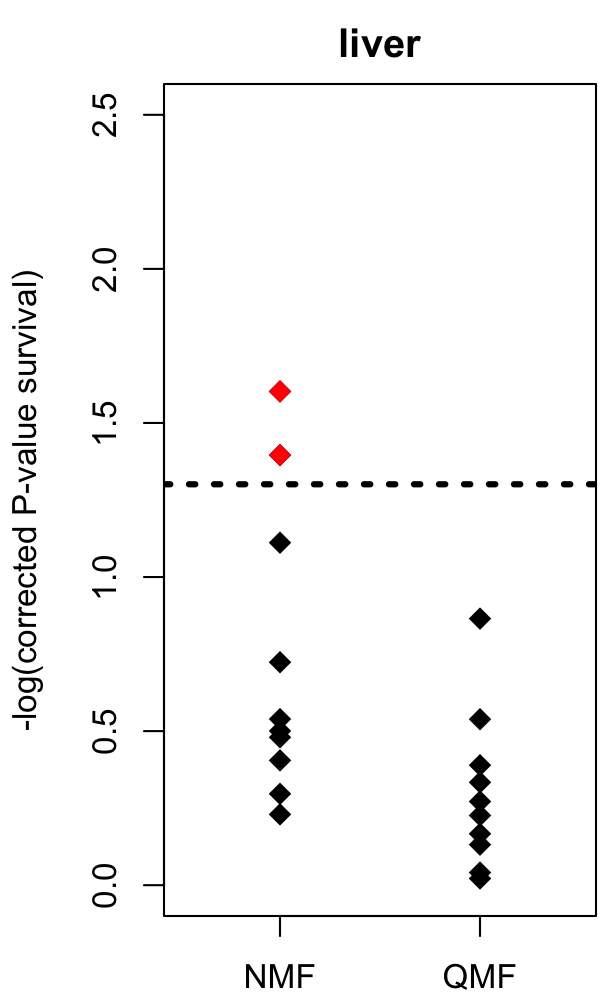}
    \includegraphics[height=6cm]{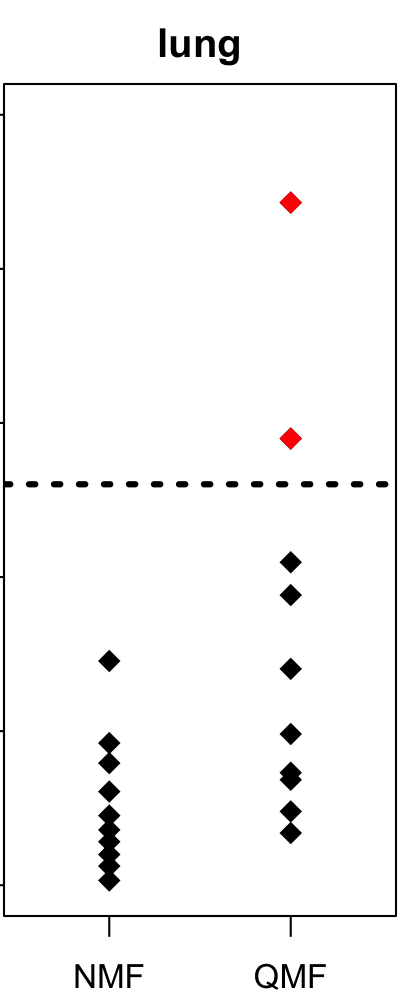}
    \includegraphics[height=6cm]{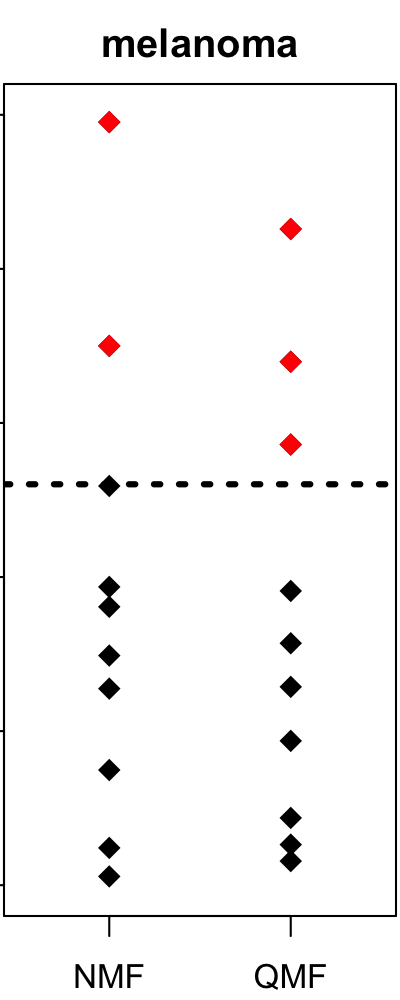}
    \includegraphics[height=6cm]{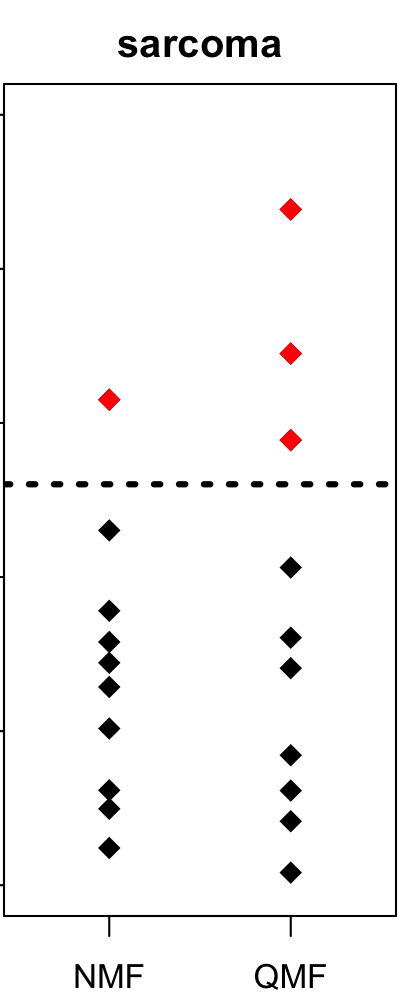}
    \caption{Identification of factors predictive of survival in cancers. For each method (NMF or QMF) and each cancer type, the dots show the negative base-10 logarithm of the Bonferroni-corrected P-values associating each of the 10 factors to survival (Cox regression-based survival analysis). The dot lines correspond to a corrected p-value of 0.05. Red dots indicate factors significantly associated with survival.}
    \label{fig:survival_pval}
\end{figure*}
To further explore the biological relevance of the factorization found by QMF, we follow the protocol of \citet{Cantini2020.01.14.905760} and compute the significance of association between each of the top 10 factors and survival, using a Cox regression-based survival analysis. For this experiment, we therefore compute a factorization in rank 10 with NMF and QMF after log-transforming the expression and methylation matrices, and keeping the top 6,000 genes with largest standard deviation in each data type. We train each model for 1,000 epochs, with a batch size of 64 and learning rate of 0.001. For QMF, we set the number of target quantiles to $m=16$, and the regularization factor to $0.001$. Figure~\ref{fig:survival_pval} summarizes the results, showing for each cancer the negative base-10 logarithm of the Bonferroni-corrected P-values associating each of the 10 factors to survival, for both NMF and QMF. Red dots indicate factors which are significantly associated with survival (i.e., p-value<0.05 after Bonferroni correction). Out of 9 cancers, we see that QMF finds more factors associated with survival than NMF in 4 cases (breast, lung, melanoma and sarcoma), while NMF outperforms QMF in one case (liver), suggesting that the factorization found by QMF is not only more accurate in terms of KL loss than the one found by NMF, but also generally more biologically relevant. Interestingly, comparing these results with the benchmark of \citet{Cantini2020.01.14.905760} which compares 8 different methods to extract relevant factors from genomic data (not necessarily based on matrix factorization), QMF matches or outperforms the best method of the benchmark in 4 cancers out of 9 (breast: 2 significant factors for QMF vs 2 for JIVE; colon: 0 for QMF vs 0 for all methods; lung: 2 for QMF vs 0 for all methods; sarcoma: 3 for QMF vs 2 for RGCCA, MCIA, Sckit and JIVE). The fact that QMF is the only method to find factors significantly associated with survival in lung cancers is particularly promising.

\section*{Conclusion.} 
We have proposed in this work an extension of low-rank matrix factorization. Our model posits that matrix factorization can be carried out, while still being reconstructive, using quantile normalization operators. These proposals are grounded on several extensions of soft-ranking and sorting operators using OT laid out in $\S2$. Our models in $\S3$ do not assume relations between the several $\bq_i$ and levels $\bb_i$, nor regularize them: this is an important future research direction. Finally, our experiments suggest that despite the non-convexity of both QMFQ and QMF, out-of-the-shelf minimizers with minimal parameter tuning provided consistently far better results than vanilla NMF. As with matrix factorization, the non-convexity of this model seems well behaved, likely to be facilitated by smoothing $\varepsilon$.
\bibliography{references.bib}
\bibliographystyle{icml2020}
\appendix
\appendix
\section{Proofs}
\subsection{Proof of Proposition~2.1}
We denote by $\Plp$ the matrix $\diag(\bu_\ell) K \diag(\bv_\ell)$ and $\Plm=\diag(\bu_{\ell-1}) K \diag(\bv_\ell)$. Recall that $\Plp\ones_m =\ba$ whereas $(\Plm)^T\ones_n =\bb$
For convenience, we can assume that the array $\bx$ is sorted in non-decreasing order and that the entries of $\bx$ are distinct.
The first assumption is without loss of generality, since applying a permutation to the entries of $\bx$ and $\ba$ has the effect of applying the same permutation to the vectors $\KR_\varepsilon$ and $\widetilde{T}_\varepsilon$.
The latter assumption can be accomplished by infinitesimally perturbing the entries of $\bx$ and using the fact that $\KR_\varepsilon$, $\Kquantile_\varepsilon$, and $\widetilde{T}_\varepsilon$ are all continuous functions of $\bx$.

Under these assumptions, it suffices to prove that the vectors $\opSquantile{\ba,\bx}{\bb,\by}$, $\opSR{\ba, \bx}{\bb, \by}$, and $\opSquan{\ba,\bx}{\by}{\varepsilon,\bb,\bq}$ are nondecreasing.
These three claims follow from the following monotonicity property of $\Plp$ and $\Plm$.

\begin{lem}\label{lem:stoch_dom}
For any $0 \leq k \leq m$, the sum of the last $k$ columns of $\diag(\ba)^{-1} \Plp$ is a vector whose entries are non-decreasing.
Similarly, for any $0 \leq k \leq n$, the sum of the last $k$ rows of $\diag(\bb)^{-1} \Plm$ is a vector whose entries are non-decreasing.
\end{lem}

Let us first see how this implies the proposition.
Let $M$ be any matrix each of whose rows sums to $1$ and such that the sum of its last $k$ columns is a non-decreasing vector.
Under these conditions, if $\mathbf{w}$ is a non-decreasing vector, then $M \mathbf{w}$ is non-decreasing.
Indeed, if we denote by $M_j$ the $j$th column of $M$, we can write
\begin{align*}
        M \mathbf{w} & = \sum_j M_j w_j = \sum_j M_j \left(w_1 + \sum_{1 \leq J < j} w_{J+1} - w_{J} \right) \\
        & = w_1 \sum_{j} M_j + \sum_{J} (w_{J+1} - w_J) \sum_{j > J} M_j\,.
\end{align*}
By assumption, $\sum_j M_j = \mathbf{1}$, the all-ones vector, and $\sum_{j > J} M_j$ is a non-decreasing vector. Since $\mathbf{w}$ is non-decreasing, $w_{J+1} - w_J$ is non-negative for each $J$.
We obtain that $M \mathbf{w}$ is the sum of a constant vector and a non-negative linear combination of non-decreasing vectors, and is therefore non-decreasing.

Applying this argument to $\diag(\bb)^{-1} (\Plm)^T$ and the non-decreasing vector $\bx$ gives the first claim on the vector of sorted values,
whereas applying the same argument to $\diag(\ba)^{-1} \Plp$ and the non-decreasing vectors $\overline{\bb}$ and $\bq$ gives the second and third claims.

All that remains is to prove the lemma.
\begin{proof}[Proof of Lemma~\ref{lem:stoch_dom}]
We prove only the the first claim, since the second follows upon taking transposes and interchanging $(\ba, \bx)$ and $(\bb, \by)$.
Write $M = \diag(\ba)^{-1} \Plp$.
Writing $M_j$ for the $j$th column of $M$, our goal is to show that $\sum_{j > J} M_j$ is a non-decreasing vector for any $J$.
Fix $i < i'$.
We first note that $j \mapsto r(j) \eqdef \frac{M_{ij}}{M_{i'j}}$ is non-increasing.
Indeed, for $j < j'$, we have $r(j)/r(j') = \frac{M_{ij}M_{i'j'}}{M_{i'j}M_{i j'}} \geq 1$ by Lemma~\ref{lem:non-dec}.
Therefore, for any $i < i'$, we have
\begin{align*}
    \sum_{j \leq J} M_{i'j}\sum_{j > J} M_{i j} & =  \sum_{j \leq J} r(j)^{-1} M_{ij}\sum_{j > J} r(j) M_{i' j} \\
    & \geq (r(m - k)^{-1}  \sum_{j \leq J} M_{ij}) (r(m-k) \sum_{j > J} M_{i'j}) \\
    & = \sum_{j \leq J} M_{ij} \sum_{j > J} M_{i' j}\,. 
\end{align*}
Recall that each row of $M$ sums to $1$.
Adding $\sum_{j > J} M_{ij} \sum_{j > J} M_{i'j}$ to both sides of the above inequality therefore yields
\begin{equation*}
    \sum_{j > J} M_{i j} \leq \sum_{j > J} M_{i' j}\,.
\end{equation*}
Since this argument holds for any $i < i'$, the vector $\sum_{j > J} M_j$ is non-decreasing, as claimed.
\end{proof}
\begin{lem}\label{lem:non-dec}
If $c$ is submodular and $\bx$ and $\by$ are non-decreasing, then for any $\ell \geq 0$ the matrix $M \eqdef \diag(\ba)^{-1} \Plp$ satisfies $M_{ij} M_{i'j'}/M_{i'j}M_{ij'} \geq 1$ for all $i \leq i'$, $j \leq j'$.
\end{lem}
\begin{proof}
By the definition of $\Plp$, we can write $M = \diag(\ba)^{-1} \diag(\bu_\ell)K\diag(\bv_\ell)$, so
\begin{align*}
    \frac{M_{ij}M_{i'j'}}{M_{i'j}M_{ij'}} & = \frac{a_i^{-1} a_{i'}^{-1} (u_\ell)_i(u_\ell)_{i'}(v_\ell)_j(v_\ell)_{j'} K_{ij} K_{i'j'}}{a_i^{-1} a_{i'}^{-1}(u_\ell)_i(u_\ell)_{i'}(v_\ell)_j(v_\ell)_{j'} K_{i'j} K_{ij'}} \\
    & = \frac{K_{ij} K_{i'j'}}{K_{i'j} K_{ij'}} \\
    & = e^{\frac{1}{\varepsilon}\left(c(x_{i'}, y_{j})+c(x_{i}, y_{j'})- c(x_i, y_j) - c(x_{i'}, y_{j'}) \right)} \\
    & = \exp\left(- \frac{1}{\varepsilon}\int_{x_i}^{x_{i'}} \int_{y_j}^{y_{j'}} \frac{\partial^2 c}{\partial x \partial y} \, \mathrm{d}y \mathrm{d}x\right) \\
    & \geq 1\,,
\end{align*}
where the last inequality follows from the assumption that $c$ is submodular.
\end{proof}

\subsection{Computing the Jacobians}

For $\bz\in\R^{n+m}$ we write $\bz_f\in\R^{n}$ (resp. $\bz_g\in\R^{m}$) for the subvector of $\bz$ with the first $n$ (resp. the last $m$) entries of $\bz$, i.e., $\bz = (\bz_f^T , \bz_g^T)^T$. Let $\Pi:\R^{n+m} \rightarrow \R^{n\times m}$ be the linear mapping defined for any $\bz\in\R^{n+m}$ by $\Pi \bz = - (\bz_f \ones_m^T + \ones_n \bz_g^T)$. For any vector $u\in\R^d$, we denote by $\diag(u)$ the $d\times d$ diagonal matrix with diagonal equal to $u$.

We define for $\bx\in\R^{n}$ and $\bz\in\R^{n+m}$ the function
$$\tau: (\bx,\bz) \mapsto \begin{bmatrix}\mine(C(\bx) + \Pi \bz) + \varepsilon \log \ba \\\mine(C(\bx)^T + (\Pi \bz)^T) + \varepsilon \log \bb\end{bmatrix}\,,$$
where $C(\bx) = [ c(x_i,y_j)]_{ij} \in \R^{n \times m}$, and for any $A\in \R^{n\times m}$, $\mine(A) = -\varepsilon \log (e^{-A/\varepsilon} \ones_m)$.

If we denote by $\bz(\bx) = (\bbf(\bx)^T,\bg(\bx)^T)^T$ the output of the Sinkhorn iterations upon convergence, then it holds that $\tau(\bx,\bz(\bx)) = 0$. $\tau$ being continuously differentiable, the implicit function theorem tells us that if the Jacobian $J_\bz \tau(\bx,\bz(\bx))$ is invertible, then there exists an open neighborhood of $\bx$ where $\bx \mapsto \bz(\bx)$ is invertible and its Jacobian satisfies $J_\bx \bz(\bx) = - J_\bz \tau(\bx,\bz(\bx))^{-1} J_\bx \tau(\bx,\bz(\bx))$. Let us therefore compute these terms.

In order to compute $- J_\bz \tau(\bx,\bz)^{-1}$, we first observe that for any $H\in\R^{n\times m}$,
$$[J_A\mine(A)](H) = \frac{(e^{-A/\varepsilon}\circ H)\ones_m}{e^{-A/\varepsilon}\ones_m},$$
therefore, for any $\delta\in\R^{n+m}$,
 $$[J_\bz\mine(C(\bx) + \Pi \bz)] (\delta) =  \frac{(M\circ \Pi \delta)\ones_m}{M\ones_m},$$
 where we write for convenience $$ M = e^{-\tfrac{C(\bx) + \Pi \bz}{\varepsilon}}.$$
Notice now that
$$ M\circ \Pi \delta = - M \circ (\delta_f \ones^T_m + \ones_n\delta_g^T) = -\diag(\delta_f)M - M\diag(\delta_g)\,,$$
therefore
$$ (M\circ \Pi \delta)\ones_m = -\delta_f\circ (M\ones_m) - M \delta_g\,,$$
from which we obtain
\begin{multline*} [J_\bz\mine(C(\bx) + \Pi \bz)](\delta) \\= - \frac{\delta_f\circ (M\ones_m) + M \delta_g}{M\ones_m}= - \delta_f -\frac{M \delta_g}{M\ones_m}\,.\end{multline*}
Similarly, we obtain
$$ [J_\bz\mine(C^T(\bx) + (\Pi \bz)^T)](\delta) = - \delta_g -\frac{M^T \delta_f}{M^T\ones_n}\,.$$
Wrapping up, we finally obtain that
$$ [J_\bz \tau(\bx,\bz)](\delta) = - \begin{bmatrix} \delta_f + \frac{M \delta_g}{M\ones_m} \\  \frac{M^T \delta_f}{M^T\ones_n} + \delta_g\end{bmatrix}\,,$$
and therefore, writing $M_1= \diag(1/{M\ones_m})M$ and $M_2=\diag(1/{M^T\ones_n})M^T$:
$$ - J_\bz \tau(\bx,\bz) = \begin{bmatrix} I_n &M_1 \\ M_2 & I_m  \\ \end{bmatrix} \,.$$
Using matrix inversion with the Schur complement, we finally get
\begin{equation}\label{eq:jz}
   - J_\bz \tau(\bx,\bz)^{-1} = \begin{bmatrix} I_n + M_1 S^{-1} M_2 & - M_1 S^{-1} \\ - S^{-1} M_2 & S^{-1}  \\ \end{bmatrix} \,,
\end{equation}
where $S = I_m-M_2 M_1$.

To compute $J_\bx \tau(\bx,\bz)$, we first observe that for any $\delta\in\R^n$,
$$
[J_\bx \tau(\bx,\bz)] (\delta) = \begin{bmatrix}[J_A \mine(C(\bx)+\Pi \bz)] \left( [J_\bx C(\bx)](\delta) \right) \\ [J_A \mine(C^T(\bx)+(\Pi \bz)^T)] \left( [J_\bx C^T(\bx)] (\delta) \right) \end{bmatrix} \,.
$$
Here, $[J_\bx C(\bx)](\delta) = \diag(\delta) \Delta$ and $[J_\bx C^T(\bx)] (\delta)= \Delta^T\diag(\delta),$
where $\Delta = [c'(x_i,y_j)]_{i,j}$. Therefore, using again the notation $M_1$ and $M_2$ , one has
\begin{equation}\label{eq:jx}
    [J_\bx \tau(\bx,\bz)] (\delta) = \begin{bmatrix} (M_1 \circ \diag(\delta) \Delta)\ones_m \\ (M_2 \circ \Delta^T\diag(\delta))\ones_n\end{bmatrix}= \begin{bmatrix} \delta\circ (M_1\circ\Delta)\ones_m \\ (M_2 \circ \Delta^T)\delta\end{bmatrix} \,.
\end{equation}

Combining (\ref{eq:jz}) and (\ref{eq:jx}), we finally get from the implicit function theorem that $[J_\bx \bz(\bx)](\delta)$ is equal to:
$$
    \begin{bmatrix}
    \left( I_n + M_1 S^{-1} M_2 \right) \left( \delta\circ (M_1\circ\Delta)\ones_m\right) - M_1 S^{-1} (M_2 \circ \Delta^T)\delta \\
    S^{-1}\left( -M_2 ( \delta\circ (M_1\circ\Delta)\ones_m) + (M_2 \circ \Delta^T)\delta \right) \\
    \end{bmatrix} \,.
$$

At this point, we should notice that the above derivation is only valid is the Jacobian $J_\bz \tau (\bx,\bz(\bx))$ is invertible. However, on easily see that for any $(\bx,\bz)\in\R^n\times \R^{n+m}$, $\tau(\bx,\bz) = \tau(\bx,\bz+\lambda \bz_0)$ with $\bz_0 = (\ones_n^T,\ones_m^T)^T$ and $\lambda>0$; and simultaneously, the $n+m$ equality in $\tau(\bx,\bz)$ are redundant, since as soon as $n+m-1$ of them are satisfied then they are all satisfied. This implies that $J_\bz \tau$ is nowhere invertible. In order to make it invertible, we can just remove the first dimension in the definition of $\tau(\bx,\bz)$, and simultaneously constrain the first coordinate of $z$ to be $0$. One can easily check that in that case, all the computations above remain valid after removing the first row/column of each matrix vector of dimension $n$.

\section{Additional experiments}
\subsection{Simulations}
In this section we provide more experimental results for the ``larger experiment'' simulated problem described in the main text, where we factorize a matrix with dimensions $d=500$, $n=256$, $k=10$, modified by a ground truth quantile normalization and corrupted by truncated Gaussian noise. Figure~\ref{fig:my_label4} showed the performance during training of NMF, QMQF and QMF with different batch size for a learning rate equal to 0.01, and $m=16$ quantiles.

\begin{figure*}
    \centering
    \includegraphics[width=\textwidth]{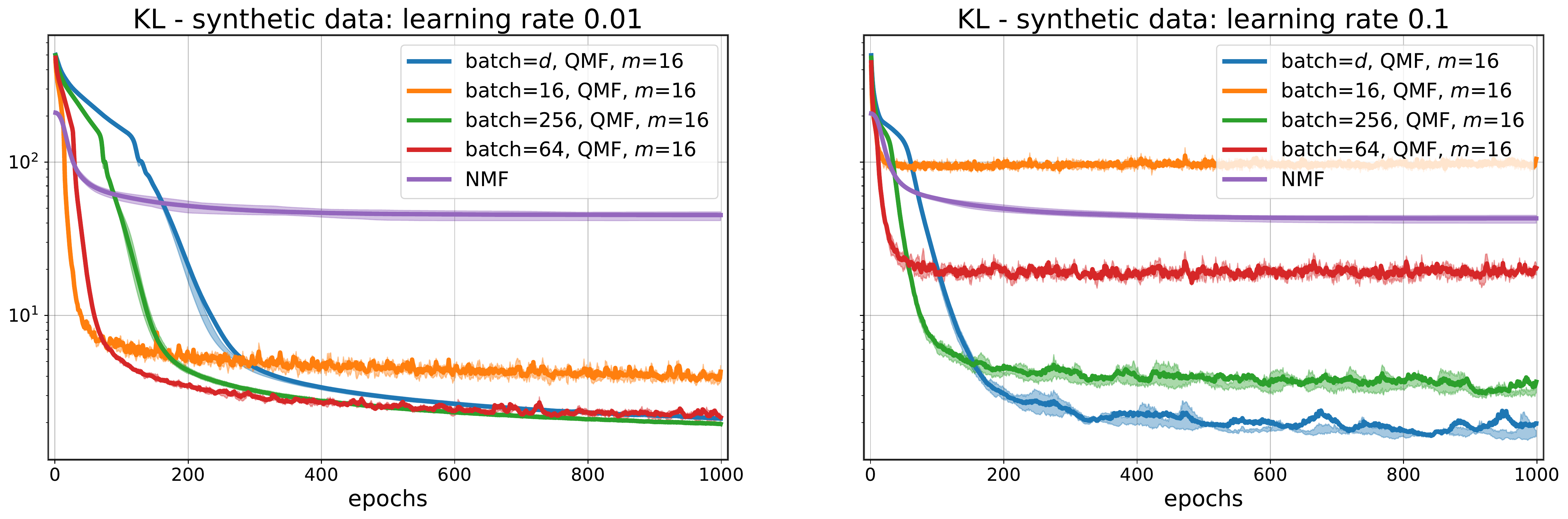}
    \caption{Sensitivity of QMF to learning rate. The setup here is identical to that of Figure 4 in the paper: we consider a synthetic model with additive censored Gaussian noise. We show the results of different methods for a learning rate equal to 0.01 \textit{(left)} or 0.1 \textit{(right)}.}
    \label{fig:1}
\end{figure*}

We first assess the influence of the learning rate. In Figure~\ref{fig:1}, we plot the performance during training of NMF and QMF with various batch size with learning rate 0.01 (left, identical to Figure~\ref{fig:my_label4}), and a larger learning rate 0.1 (right). While NMF does not seem to be influenced by the learning rate in this case, we see that the performance of QMF degrades when the learning rate is too large, particularly for small batch sizes, as expected. Overall, this confirms that taking 0.01 allows QMF to converge to a good solution, at least when the batch size is at least 64.

\begin{figure}
    \centering
    \includegraphics[width=0.5\textwidth]{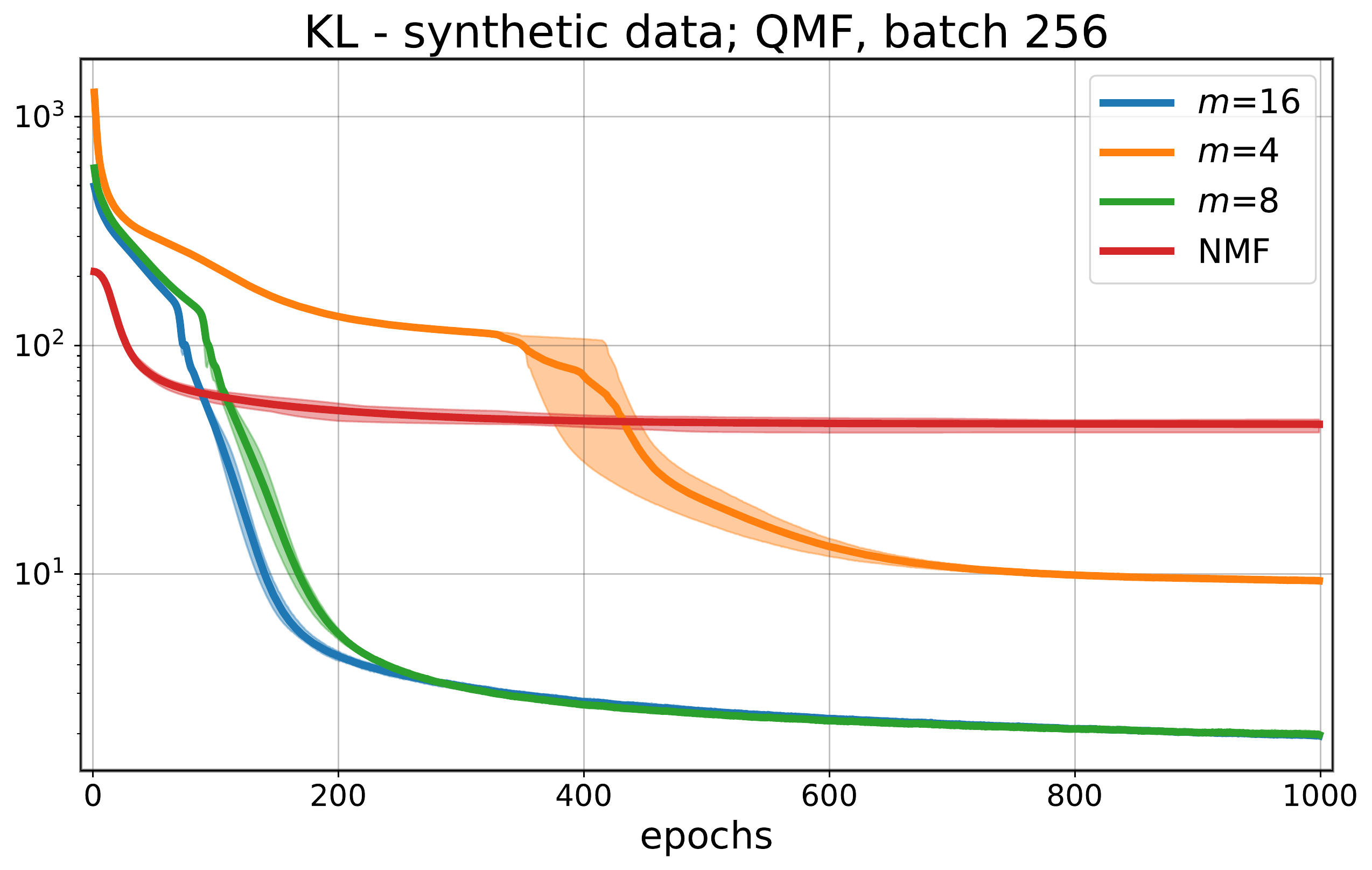}
    \caption{Sensitivity of QMF to the number of target quantiles: we have observed that setting $m$ to a number larger than $8$ is usually sufficient to obtain good results. Here again the setup is identical to that of Figure 4, with a learning rate set to 0.01}
    \label{fig:2}
\end{figure}

Second, we discuss the impact of $m$, the number of quantile levels. Figure~\ref{fig:2} shows the training error of QMF when the learning rate is fixed to 0.01, and we vary $m$ among 4, 8 and 16. We see that $m=4$ leads to a suboptimal approximation compared to $m=8$ or $m=16$, suggesting that $m$ should be large enough to model the quantile transformation. On the other hand, the fact that $m=16$ is not better than $m=8$ (while the ground truth quantile transformation is obtained with $m=256$ quantile levels) suggests that a relatively small number of quantile levels is enough to approximate a complex transform, in that case.

\begin{figure*}
    \centering
    \includegraphics[width=\textwidth]{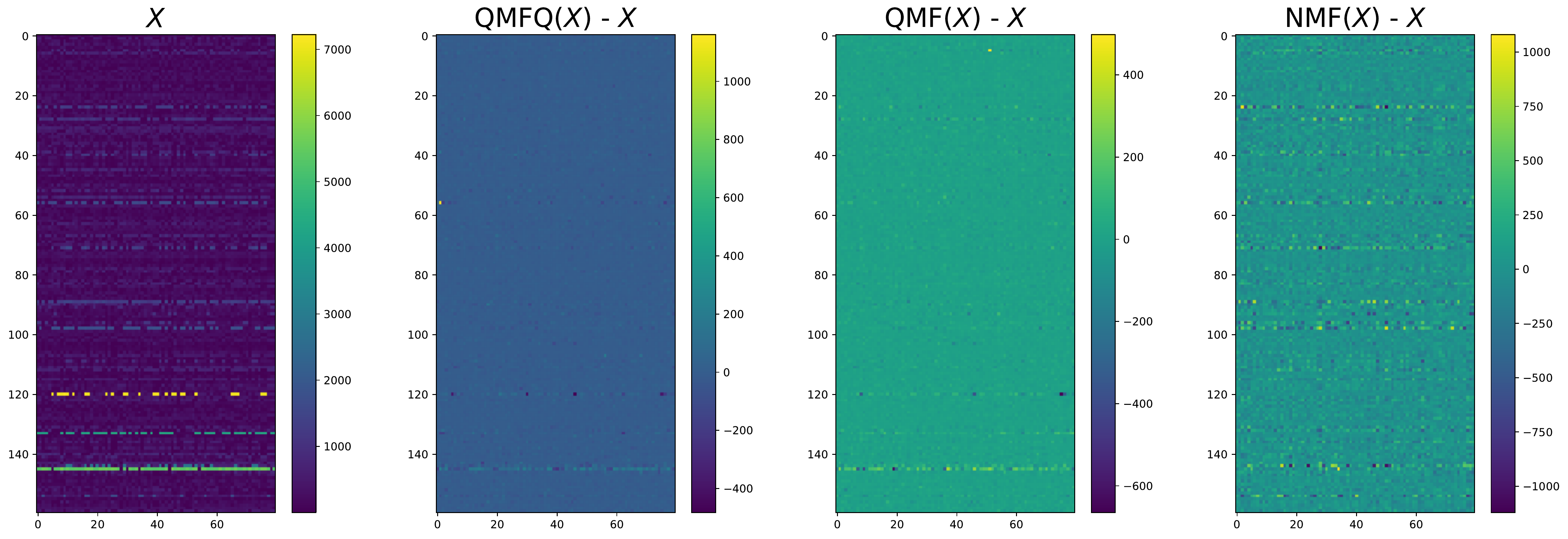}
    \caption{Example of data matrix on the left, along reconstruction errors of all 3 approaches considered here, QMFQ, QMF and NMF.}
    \label{fig:lr}
\end{figure*}

Figure~\ref{fig:lr} illustrates the different behaviors of NMF, QMF and QMFQ on a simple matrix $X$ (simulated according to the ``toy illustration'', with $d=160$, $n=80$, $k=8$, see main text for details), where we see strong row-wise patterns due to different quantile transformations applied rowwise. We see in particular the that residuals after matrix approximation by NMF have still strong rowwise patterns, and overall larger values than those after QMF and QMFQ approximation.

\begin{figure*}
    \centering
    \includegraphics[width=\textwidth]{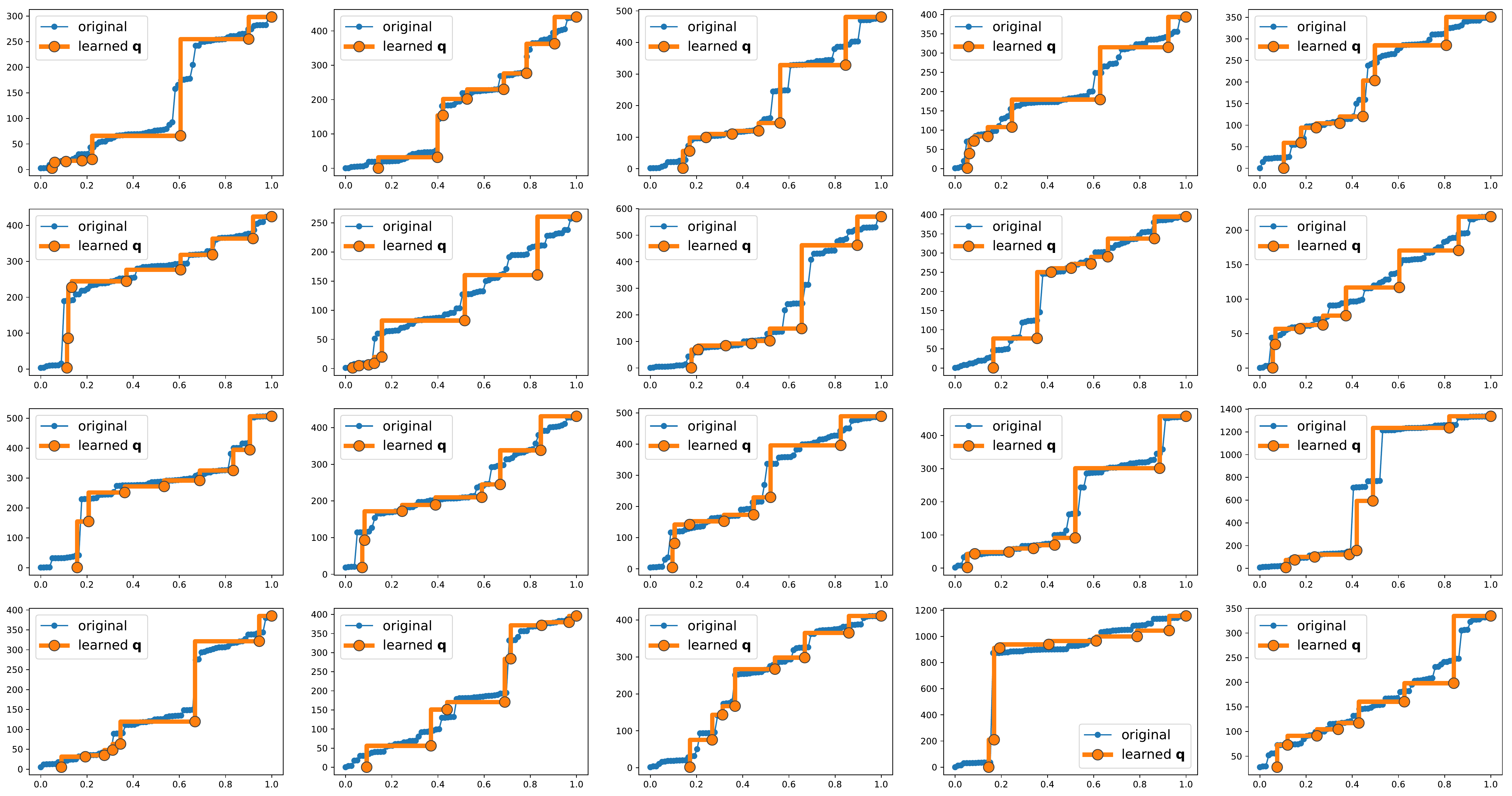}
    \caption{Reconstruction of quantile distributions for QMF in the synthetic + noise setting of Fig. 4 in the paper for the first 20 features.}
    \label{fig:4}
\end{figure*}

\begin{figure*}
    \centering
    \includegraphics[width=\textwidth]{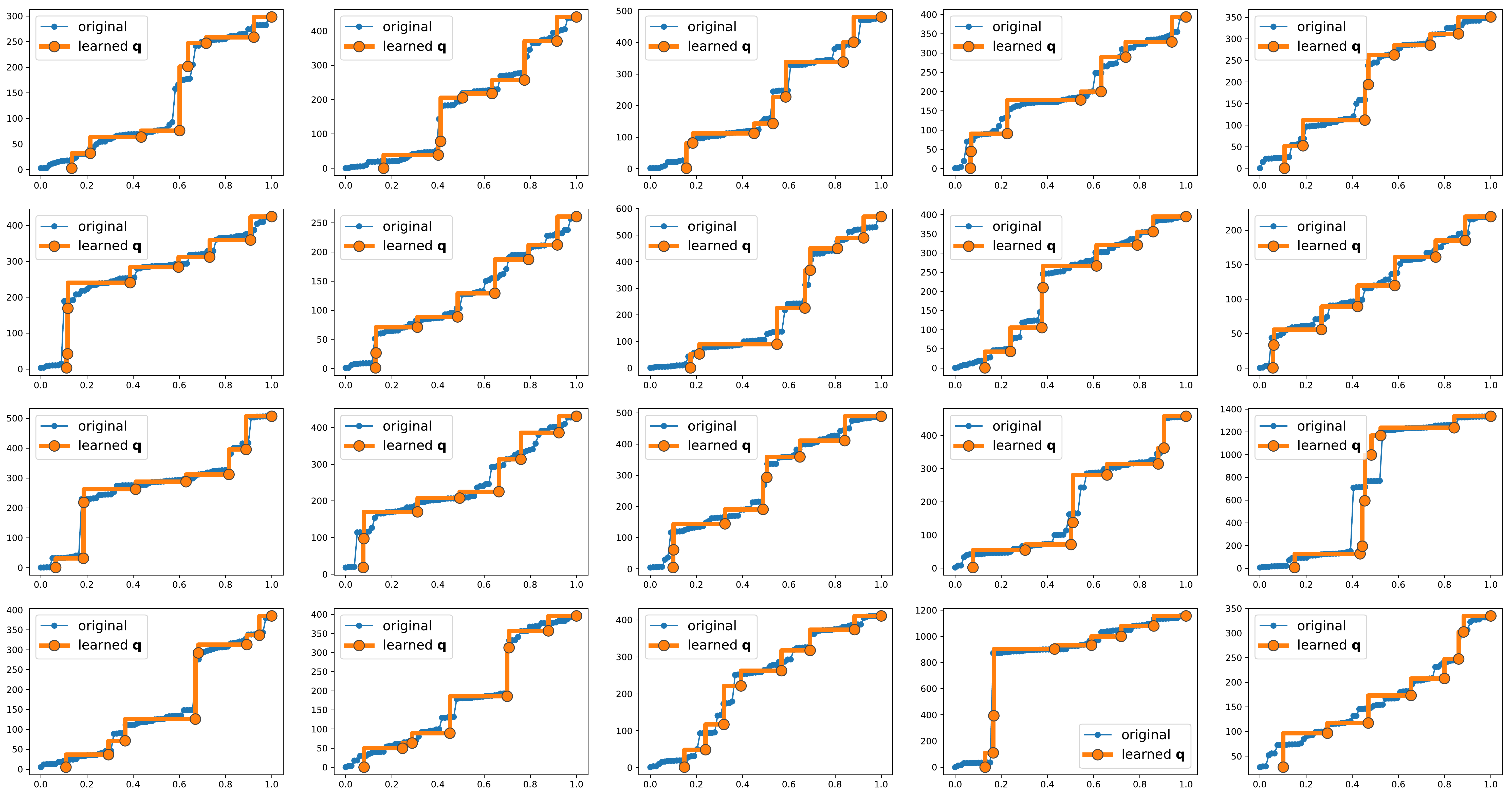}
    \caption{Reconstruction of quantile distributions for QMFQ in the synthetic + noise setting of Fig. 4 in the paper for the first 20 features.}
    \label{fig:5}
\end{figure*}

In Figures \ref{fig:4} and \ref{fig:5}, finally, we compare the quantile transforms inferred by QMF and QMFQ, respectively, on the ``larger experiment'' with the parameters of Figure~\ref{fig:my_label4}. Each figure shows the quantile functions inferred for the first 20 features (out of a total of $d=500$ features). While the reconstructed quantiles are generally very good approximations of the ground truth (in blue), we see a few cases where QMFQ (a more costly option) recovers slightly better the ground truth quantile function than QMF. In particular, it seems that QMF sometimes allocates its budget of quantile values not optimally (e.g. lower left plot of Figure \ref{fig:4}) whereas QMFQ does a better job in Figure~\ref{fig:5}. It would be interesting to better understand why we see this behavior.

\subsection{Genomics}
\begin{figure*}
    \centering
    \includegraphics[width=\textwidth]{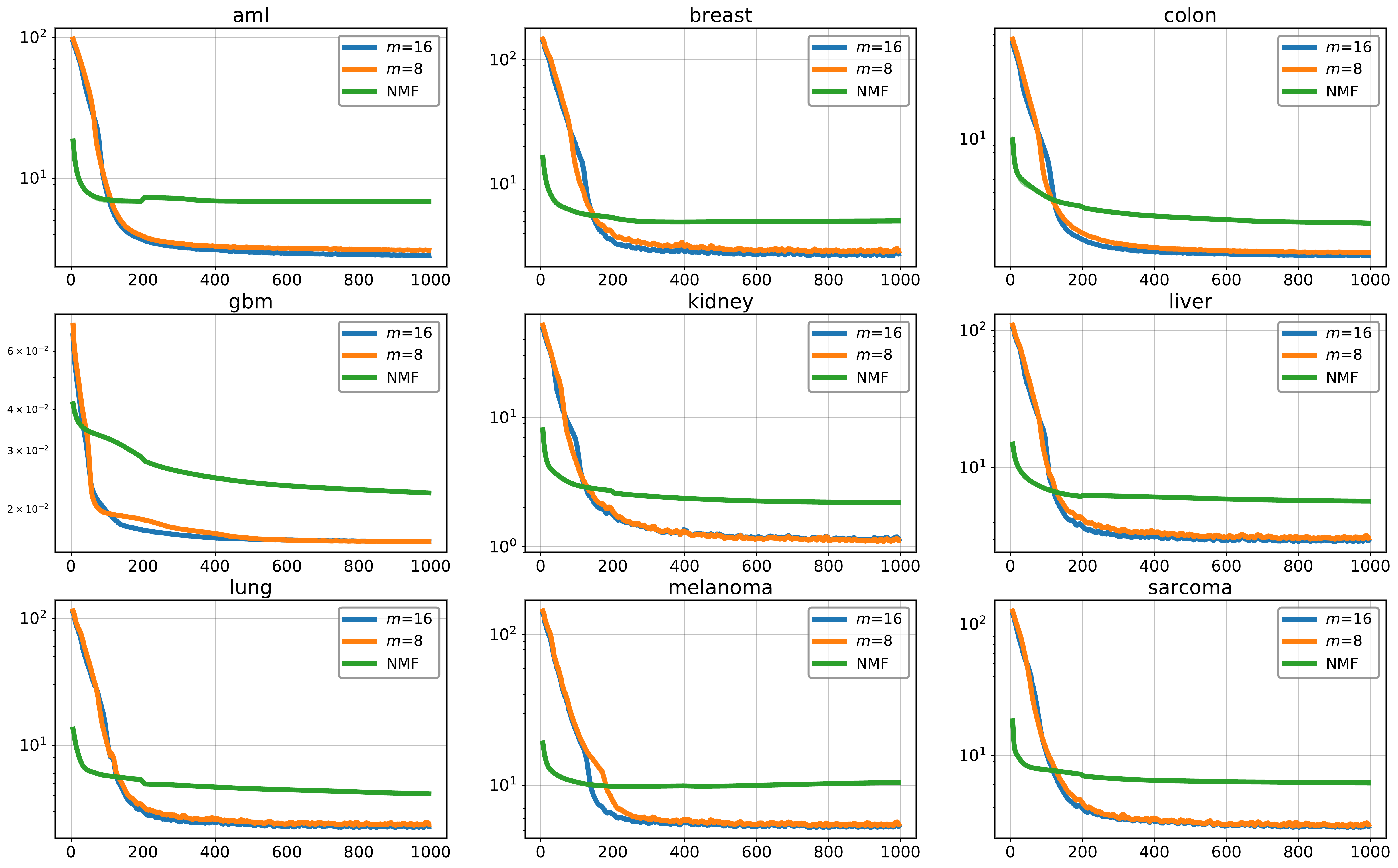}
    \caption{Decrease of Kullback-Leibler divergence on the 9 genomics datasets, using QMF with a batch size of 64 and a learning rate of 0.001 with different number of quantiles $m$.}
    \label{fig:lr_genomics}
\end{figure*}

In this section we provide additional experimental results regarding the use of QMF for cancer genomics data integration. In particular, to assess the influence of the number of quantile values $m$, we show in Figure~\ref{fig:lr_genomics} the decrease in KL loss during optimization, on the 9 cancer data sets, for NMF and for QMF with $m=8$ or $16$ target quantiles. The loss tends to decrease initially faster with NMF, but after about 100 iterations QMF reaches lower loss values than NMF consistently across all cancers and converges to lower values. We do not see any important difference between $m=8$ and $m=16$.

\end{document}